\documentclass[twoside,11pt]{article}

\usepackage[utf8]{inputenc} 
\usepackage[T1]{fontenc}    
\usepackage{url}            
\usepackage{booktabs}       
\usepackage{amsfonts}       
\usepackage{nicefrac}       
\usepackage{xcolor}         
\usepackage{graphicx}
\usepackage{subfigure}
\usepackage[ruled,vlined]{algorithm2e}
\usepackage{amsmath}
\usepackage{amssymb}
\usepackage{mathtools}
\usepackage{amsthm}
\usepackage{array}
\usepackage{multirow}
\usepackage{bm}
\usepackage{BOONDOX-cal}

\newtheorem{assumption}{Assumption}
\allowdisplaybreaks[4]

\usepackage{jmlr2e}

\usepackage{lastpage}
\jmlrheading{23}{2023}{1-\pageref{LastPage}}{1/21; Revised 5/22}{9/22}{21-0000}{Feifei Wang, Huiyun Tang, and Yang Li}

\ShortHeadings{Factor-Assisted Federated Learning}{Wang et al.}
\firstpageno{1}

\begin{document}

\title{Factor-Assisted Federated Learning for Personalized Optimization with Heterogeneous Data}

\author{\name Feifei Wang \email feifei.wang@ruc.edu.cn \\
	\addr Center for Applied Statistics\\
	Renmin University of China\\
	Beijing, 100872, China
	\AND
	\name Huiyun Tang \email 2020103671@ruc.duc.cn \\
	\addr School of Statistics\\
	Renmin University of China\\
	Beijing, 100872, China
	\AND
	\name Yang Li \email yang.li@ruc.edu.cn \\
	\addr Center for Applied Statistics\\
	Renmin University of China\\
	Beijing, 100872, China}

\editor{My editor}

\maketitle

\begin{abstract}
Federated learning is an emerging distributed machine learning framework aiming at protecting data privacy.
	Data heterogeneity is one of the core challenges in federated learning, which could severely degrade the convergence rate and prediction performance of deep neural networks. To address this issue, we develop a novel personalized federated learning framework for heterogeneous data, which we refer to as FedSplit. This modeling framework is motivated by the finding that, data in different clients contain both common knowledge and personalized knowledge. Then the hidden elements in each neural layer can be split into the shared and personalized groups. With this decomposition, a novel objective function is established and optimized. We demonstrate FedSplit enjoyers a faster convergence speed than the standard federated learning method both theoretically and empirically. The generalization bound of the FedSplit method is also studied. To practically implement the proposed method on real datasets, factor analysis is introduced to facilitate the decoupling of hidden elements. This leads to a practically implemented model for FedSplit and we further refer to as FedFac. We demonstrated by simulation studies that, using factor analysis can well recover the underlying shared/personalized decomposition. The superior prediction performance of FedFac is further verified empirically by comparison with various state-of-the-art federated learning methods on several real datasets.
\end{abstract}


\section{Introduction}\label{section1}
Federated learning (FL) is a novel distributed machine learning approach \citep{mcmahan2017communication}. The standard FL algorithm involves a set of clients with locally stored datasets. All clients collaborate to train a global model under the supervision of a central server. In this work, we focus on horizontal federated learning \citep{yang2019federated}, which means datasets on different clients share the same feature space but different in samples. In horizontal federated learning (we omit ``horizontal" for simplicity hereafter), one of the core challenges is the heterogeneity issue because data in different clients are generated separately and never mixed up \citep{li2020federated,kairouz2021advances,li2021survey}. Data heterogeneity violates the frequently used independent and identically distributed (IID) assumption in distributed optimization, which is important for the stochastic gradient descent (SGD) type algorithms \citep{mcmahan2017communication}. When modeling heterogeneous data in federated learning, the SGD algorithms could generate biased estimates and hurt the training convergence and prediction performance \citep{li2019convergence,karimireddy2020scaffold,li2020federated,Zhang_2021}. In addition, the estimation and prediction performance could also vary greatly across clients with heterogeneous data \citep{li2019fair,mohri2019agnostic} and thus hurt the generalization ability of FL models.

There exist various studies to address the heterogeneity challenge, which can be roughly divided into two streams. The first stream of studies try to extend the standard FedAvg method \citep{mcmahan2017communication} to allow for more stable performance and more robust convergence in heterogeneous settings \citep{li2019fair,mohri2019agnostic,karimireddy2020scaffold, acar2021federated}. However, these works still produce a single global model and can only deal with certain degree of heterogeneity \citep{tan2022towards}. Another stream of studies focus on personalizing the global model to be more customized to individual clients \citep{li2020federated, huang2021personalized, kairouz2021advances, tan2022towards}. As personalized models are preferred in many practical applications such as healthcare, finance, and AI services, this stream of studies have gained increasing popularity in recent years \citep{li2019fedmd, wang2019federated, ge2020fedner, PEI2022108906}. More thorough discussion for related works can be found in Section 2.

\begin{figure}[h]
	\centering
	\includegraphics[width=0.8\textwidth]{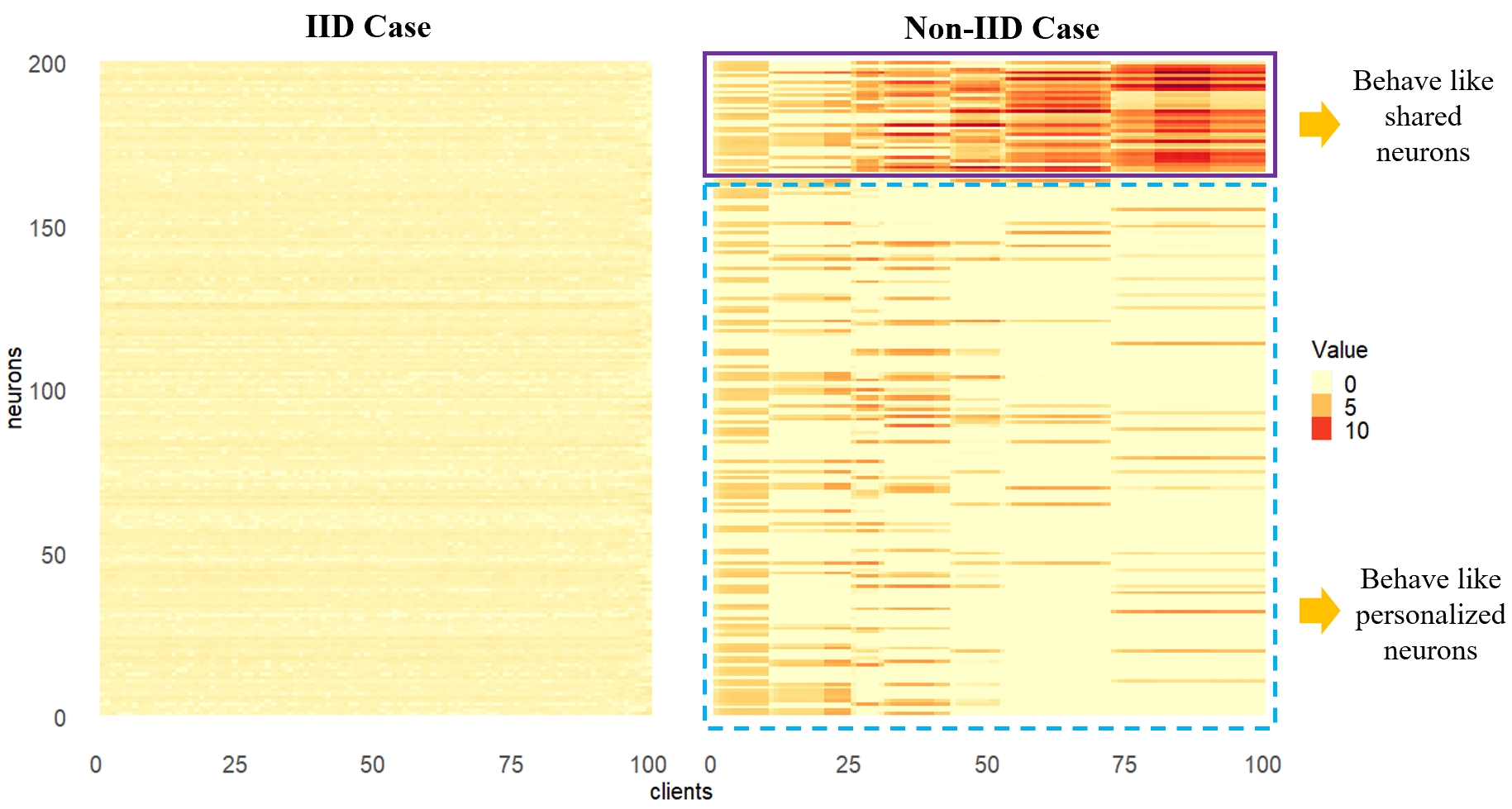}
	\caption{The heatmaps of outputs generated by each neuron in each client in the IID and non-IID cases.}
	\label{my-example}
\end{figure}

\begin{figure}[h]
	\centering
	\includegraphics[width=0.8\textwidth]{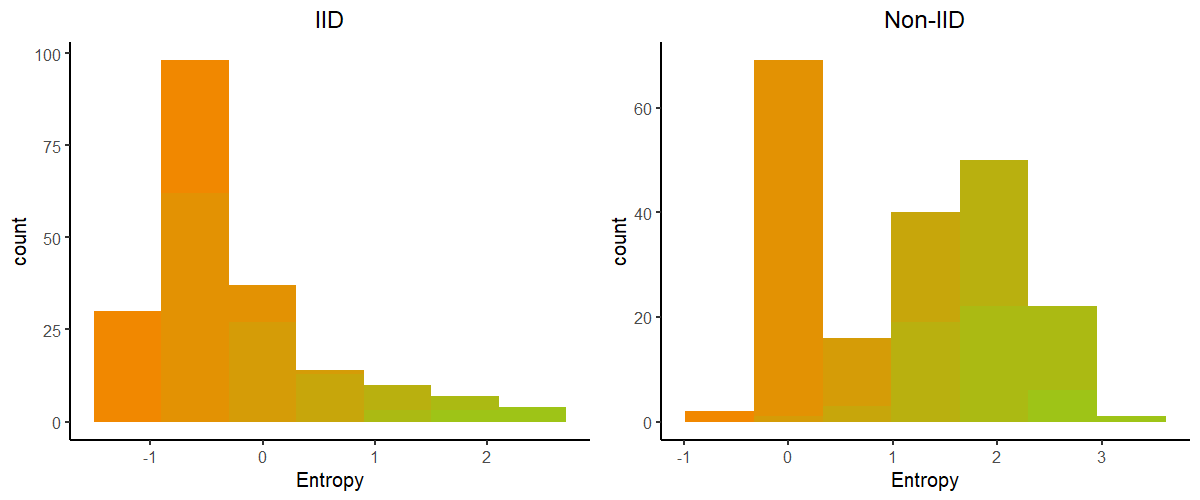}
	\caption{The histograms of entropy of outputs generated by each neuron across all clients in the IID and non-IID cases. }
	\label{example2}
\end{figure}

The past literature mainly focuses on manipulating models. However in this work, we try to utilize the characteristics of heterogeneous data itself to solve the heterogeneous problem. Our proposed method is based on one common phenomenon. That is, data in different clients contain both common knowledge and personalized knowledge. Accordingly, to model the heterogeneous data in different clients, the hidden elements in each neural layer of the deep neural networks (DNNs) might also contain the shared and personalized groups. To illustrate this idea,
we first design a heterogeneous setting using MNIST data. Specifically, we divide the dataset into ten categories by the digit labels. Each digit category is further divided into ten groups. The resulting 100 sub-datasets constitute 100 clients with heterogeneity (i.e., the non-IID case). We also consider the IID case for comparison, where the MNIST data are randomly split into 100 sub-datasets. In each case, we separately train a multi-layer perception model with one hidden layer of 200 units for digit classification for each client. Figure~\ref{my-example} shows the heatmaps of outputs from each neuron in each client.
As shown, nearly all neurons perform similarly for all clients in the IID case. However, there exist neurons learning quite varied representations across clients in the non-IID case. Specifically, the top neurons in the right panel of Figure~\ref{my-example} have large output values for most clients, which reflect \emph{common representations}. The outputs of other neurons only behave large for some specific clients, which reflect \emph{client-specific representations}. To further quantify this phenomenon, we compute the entropy of outputs generated by each neuron across all clients using the $k$ nearest-neighbor method \citep{kozachenko1987sample}. The bigger the entropy, the greater the dispersion degree. Figure~\ref{example2} compares the distributions of entropy in the IID and non-IID cases. As shown, the IID case has most entropy values near or below zero, meaning the representations of neurons across all clients in the IID case are quite concentrated. In the non-IID case, the entropy values have two peaks, one near zero and the other near two. This finding indicates that some neurons extract similar representations and behave concentrated, while some other neurons generate heterogeneous representations and behave dispersive.

\begin{figure}[h]
	\begin{center}
		\centerline{\includegraphics[width=0.6\textwidth]{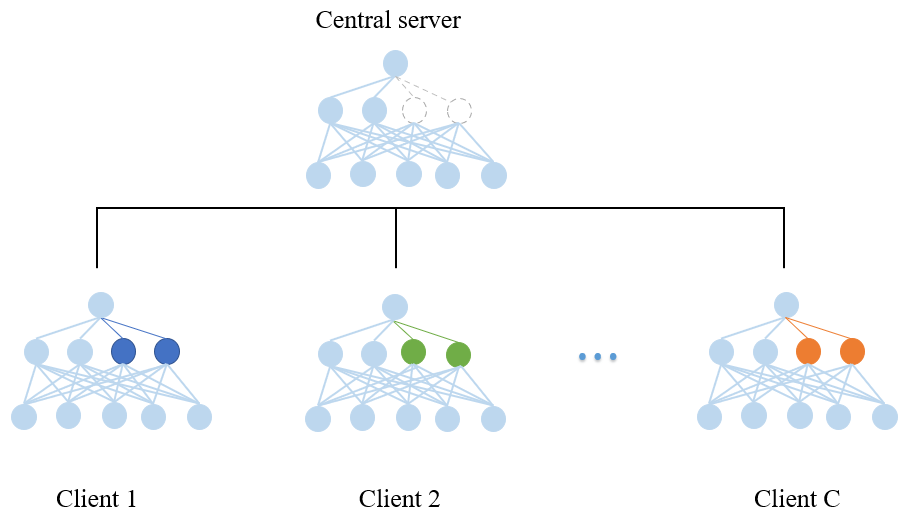}}
		\caption{The illustration of neurons decomposition in DNNs under the FedSplit framework. The light blue circles represent the shared elements, which
are updated by all clients, while circles in other colors represent the client-specific elements, which are only updated locally.}
		\label{example3}
	\end{center}
\end{figure}

The above results motivate us to decompose neurons into \emph{client-shared} ones and \emph{client-specific} ones according to their representation ability. Assume under some certain splitting method, we can split the hidden elements of each layer (i.e., channels for convolution layers or neurons for fully connected layers) into the client-shared group and the client-specific group. See Figure \ref{example3} for an illustration, where the light blue circles represent the shared hidden elements and circles in other colors represent the personalized hidden elements of each client. Then parameters corresponding to the client-shared group are updated by all clients, which are in accordance with those in the standard FedAvg method. However, the parameters corresponding to the client-specific group are not synchronized with the shared ones in the updating procedure. They are only updated locally and not averaged by the server. By this way, the resulting models are more customized for individual clients. We refer to this method as FedSplit. We provide theoretical analysis that guarantees the linear convergence and generalization property of the FedSplit method, which is trained via the parallel gradient descend algorithm.

In complicated learning scenarios with image or text, it is not easy to perform such decomposition directly on the neurons. To tackle this problem, we adopt the statistical technique of factor analysis \citep{harman1976modern} to accomplish this decomposition, which we refer to as \emph{FedFac} in the subsequent analysis. Specifically, we regard the hidden layers of DNN as higher-level representations of the raw data. Then, factor analysis is applied on the hidden elements of each layer to split them into client-shared ones and client-specific ones. Factor analysis is a classic statistical method, which can discover the inter-dependence structure among several variables. In this work, we apply factor analysis to find the inter-dependence structure among hidden elements within each layer, based on which the hidden elements would be split into the client-shared group and client-specific group. After decomposition, the client-shared elements would be updated using all client information, while the client-specific ones would only be updated for specific clients. Two specific solutions (static or dynamic) are provided to implement the decomposition of client-shared elements and client-specific elements using factor analysis. By various simulation experiments, we find both solutions can well recover the true underlying decomposition. Finally, we empirically demonstrate the benefits of FedFac in several real-world datasets when compared with various state-of-the-art FL models.



This paper is organized as follows. Section \ref{sec2} reviews the related literature. Section \ref{sec3} introduces the FedSplit methodology with decomposition of client-shared and client-specific elements. The theoretical properties of convergence and generalization ability are also provided. Section \ref{fad} discusses the implementation of FedSplit via factor analysis (i.e., the FedFac method). Its performance is verified through simulation studies. Section \ref{experiments} presents extensive experiments comparing performance of FedFac against some state-of-the-art federated learning approaches. Section \ref{section5} concludes the paper with a brief discussion.

\section{Related Work}
\label{sec2}

\subsection{Heterogeneity adjusted global models}
There are two mainstreams under this strategy.
	One is to handle the non-IID problem by modifying the standard FedAvg model \citep{mcmahan2017communication}. Some important modifications include: adopting the strategy of learning rate decay \citep{li2019convergence}, choosing server-side adaptive optimizers \citep{reddi2020adaptive}, normalizing the local model updates when making averages \citep{wang2020tackling}, and correcting the local gradient updates towards the global true optimum \citep{karimireddy2020scaffold, mitra2021linear}.
Another line of works develop new algorithms to train the global model with heterogeneity. \cite{mohri2019agnostic} regarded the distribution of the whole data in the federated system as a convex combination of multi-source distributions, and then solved the federated learning problem with min-max optimization. The generalization ability of the obtained model is lower bounded under other unfavorable distributions. \cite{li2020federated} introduced a proximal term to the local objective function, so as to penalize the local updates which are far away from the global model. \cite{karimireddy2020scaffold} and \cite{mitra2021linear} introduced control variables to correct the updating directions of local gradients on the client side, thereby reducing the variance between local gradients and enabling the solutions of local models to converge towards the optimal solution of the global model.

These heterogeneity adjusted global models alleviate the challenges brought by data heterogeneity. However there still exist some shortcomings. On one hand, these methods are only verified to be effective under some certain ``bounded-non-IID" conditions. On the other hand, they inherit the limitations of the resulted global model.
	First, the global model is unable to capture individual differentiated information of each client, harming the inference or prediction performance when deployed to clients. Second, it requires all participating clients to reach a consensus for collaborative training, which is unrealistic in practical complex IoT (Internet of Things) applications. Moreover, the global model could perform quite differently across clients, which results in unfairness in terms of the prediction performance \citep{kairouz2021advances}. Therefore some researchers have questioned global models. For example, \cite{hanzely2020federated} judged the practicality of a global model that is beyond the typical use of the clients. \cite{yu2022salvaging} suggested that for many tasks, some clients may not be able to benefit from participating in federated learning because the global shared model is not as accurate as the local models trained by themselves.

\subsection{Personalization models}
Given that a single global model cannot flexibly adapt to the diverse needs of various clients, researchers have attempted to personalize the learning process considering heterogeneity in devices, data, and models. This leads to personalized federated learning, which we call PFL for short. PFL aims to collaboratively learn personalized models for each client, thus inherently fitting the heterogeneity situation.
	PFL can be summarized into two main categories: one is to personalize the global model directly, and the other is to develop specific models for all clients through analyzing and mining the interrelationships between data or tasks across clients. The classic methods in the first category include: fine-tuning \citep{wang2019federated, collins2022fedavg,yu2022salvaging} and meta-learning \citep{jiang2019improving, fallah2020personalized}.
	The main idea is to first train a single global model in the federated learning framework, and then conduct additional training of the global model on each client to obtain a locally adaptive model. Note that the performance of the personalized model depends on the generalization ability of the global model. Then many methods under this scheme aim to improve the performance of the global model under data heterogeneity.
	The second category includes: federated multi-task learning \citep{smith2017federated, corinzia2019variational, marfoq2021federated}, interpolation of local and global models \citep{Deng2020three,mansour2020three, marfoq2022personalized}, clustered federated learning \citep{ghosh2020efficient, mansour2020three, sattler2020clustered}, and model regularization \citep{hanzely2020federated,t2020personalized}. These methods strive to utilize relevant information between clients to construct similar personalized models for related clients and thus improve the performance of customized models. See the comprehensive surveys in \cite{kairouz2021advances} and \cite{tan2022towards} for more details.

\subsection{Split-personalization models}
Among the personalization models, \emph{split-personalization} \citep{tan2022towards} is especially designed for deep neural networks.
It is an extension of split learning in the context of personalized federated learning. It is first introduced by \cite{arivazhagan2019federated}, in which a neural network is divided into the base layers and personalized layers. Then the base layers are trained by the central server. \cite{collins2021exploiting} proposed to update the base layers and personalized layers for different number of epochs. They also theoretically proved that, the shared data representations learned by the base layers could converge to the ground-truth representations. \cite{oh2022fedbabu} concentrated on training the base layers with the head randomly initialized and never updated, then the head was fine-tuned for personalization.
\cite{liang2020think} suggested local representation learning finished independently by individual clients and then the head layers trained globally. The strategy to exclude some DNN layers from averaging is also employed by \cite{li2021fedbn} to mitigate distribution shift in the feature space.
The existing split-personalization methods are mainly layer-wise. That is, they allow different layers to have different behaviours across clients, but the hidden parameters within the same layer should be aligned across clients. Compared with the previous split-personalization methods, the idea of splitting neuron-wise seems more flexible, which is inline with the emerging paradigm of \emph{split channel-wise}. In this regard, \cite{shen2022cd2} strived to personalize the FL model through
	dividing the DNN into two sub-models by splitting each layer from bottom to top, and then constrained the two sub-models to align with each other. However, they lacked of discussion about how to split channels in the FL model. \cite{alam2022fedrolex}
extracted smaller sub-models from the global model for partial training so as to fit for the heterogeneous capabilities of clients. As for decomposition, they utilized a rolling window method to select the client-specific nodes. The rolling window advances in each round and loops over all node indices in sequence progressively. By this way, they in fact did not consider the varied representation capability of nodes.

In this work, we develop a more refined decoupling strategy than the existing split-personalization methods.
	Although we also target on decoupling for the DNN structure, our proposed FedSplit method is distinct from the aforementioned methods. Specifically, we focus on developing a DNN framework that is capable of capturing the client-specific representations and common representations, so that both the representations could be incorporated into the final shared prediction model to customize the global model. Furthermore, we provide theoretical analysis that guarantees the linear convergence and generalization property of the FedSplit method trained with parallel gradient descend algorithm. However, the previous split-personalization methods usually lack theoretical guarantee. To our best knowledge, we make the first step to bridge the theoretical gap between the empirical success of personalizing FL models and splitting DNN layers vertically. Last, we propose to use factor analysis for decomposition of the two types of hidden elements that generate discrepant representations. The good performance of using factor analysis has been verified through both simulation studies and empirical experiments.

\section{The FedSplit Methodology}\label{sec3}

\subsection{Problem formulation}

Consider a FL system with $C$ clients and $N$ samples. For each client $c \in [C]$, we can observe $n_c$ observations generated according to a local distribution $\mathcal{D}_c$ over $\mathcal{X} \times \mathcal{Y}$, where $\mathcal{X} \in \mathbb{R}^{d}$ is the input domain, $\mathcal{Y} \in \mathbb{R}$ is the output domain, and $N = \sum_{c \in [C]}n_c$. Denote $D_c$ as the set of the $n_c$ observations.
For any model $\mathcal{h} \in \mathcal{H}$ with $\mathcal{h}: \mathcal{X} \mapsto \mathcal{Y}$, the loss function is defined as $\ell: \mathcal{Y} \times \mathcal{Y} \rightarrow \mathbb{R}$. Usually, $\mathcal{h}$ is a deep neural network. The goal of the standard FL method FedAvg \citep{mcmahan2017communication} is to fit a shared global model, which has good averaged performance on all clients. However, in a heterogeneous setting where the data distributions $\mathcal{D}_c$ vary across the clients, such a global model could not generalize well on individual clients. To address this issue, it is natural to fit separate model $\mathcal{h}_c \in \mathcal{H}$ for data distribution $\mathcal{D}_{c}$ on client $c$. Denote $\mathcal{L}_{\mathcal{D}_c}(\mathcal{h}_c)=\mathbb{E}_{(\boldsymbol{x}, y) \sim \mathcal{D}_{c}}[\ell(\mathcal{h}_c(\mathbf{x}), y)]$ as the true risk of $\mathcal{h}_c$ at distribution $\mathcal{D}_c$. Then the objective function for heterogeneous federated learning is:
\begin{equation}
	\forall 1\leq c \leq C, \quad \underset{\mathcal{h}_{c} \in \mathcal{H}}{\operatorname{min}} \mathcal{L}_{\mathcal{D}_{c}}\left(\mathcal{h}_{c}\right).
\end{equation}

Although the $C$ clients have different local distributions, there may exist common knowledge as well as personalized knowledge among the heterogeneous data stored in different clients; recall the motivating example discussed in Section 1. Motivated by this idea, we aim to distinguish between the two types of knowledge. Then we use the common knowledge in cooperative training of the shared global model, while use the personalized knowledge to improve the learning tasks on specific clients. Note that data are often unstructured in complicated learning scenarios. It is difficult to operate decomposition on data directly. To solve this problem, we regard the parameters associated with the hidden elements in deep neural networks as a substitute, and propose the FedSplit method.

Concretely, assume the model of research interest is a $L$-layer deep neural network with layers numbered from $0$ (input) to $L$ (output). Each layer contains $m_l, l \in [L]$ hidden elements with a Lipschitz nonlinearity function $\sigma: \mathbb{R} \rightarrow \mathbb{R}$. Denote $\mathbcal{W}$ to be the parameter set in this neural network with $p$ dimension. According to the work of \cite{jacot2018neural}, training the neural network is equivalent to linear regression using non-linear neural tangent feature space (NTFs) as the layer width approaches infinity. For each data point $\mathbf{x} \in \mathcal{X}$, the corresponding feature is given by $\phi(\mathbf{x})=\nabla_{\mathbcal{W}} \mathcal{h}(\mathbf{x};\mathbcal{W}) \in \mathbb{R}^p$. Notebly $\phi(\mathbf{x})$ could vary across clients due to the heterogeneity of local distributions. Then the feature space should contain both client-shared and client-specific information. To separate this two types of information, we focus on the parameter set $\mathbcal{W}$, which can be regarded as the affine transformation of feature $\phi(\mathbf{x})$. However, the parameter set $\mathbcal{W}$ is often very huge, it is difficult to decompose $\mathbcal{W}$ directly. Note that for DNN models, parameters of one layer are incorporated in the next layer, thus a compromise is to decompose the parameter set layer-wise. Specifically, let $\Lambda \subset \{1,\cdots, L-1\}$ denote the set of layer indices needing decomposition. For the $l$-th layer with $l \in \Lambda$, denote the parameter set for client $c$ in this layer by $\mathbcal{w}_{cl}$, which can be decomposed into two parts: (1) the parameter set $\mathbcal{w}_{l}^{\text{s}}$ shared by all clients, and (2) the personalized parameter set $\mathbcal{w}_{cl}^{\text{p}}$ for client $c$. Then we have $\mathbcal{w}_{cl}=\{\mathbcal{w}_{l}^{\text{s}},\mathbcal{w}_{cl}^{\text{p}}\}$ for $c \in [C]$ and $l \in \Lambda$. For the $j$th layer with $j \notin \Lambda$, decomposition is not needed and the parameters are shared by all clients. Then we have $\mathbcal{w}_{cj}=\mathbcal{w}_{j}^{\text{s}}$ for $c \in [C]$.
Define $\mathbcal{W}_{c}$ as the set of all parameters for client $c$, which includes both shared parameters and personalized parameters.
Then the objective function of FedSplit is
\begin{equation}
	\underset{\mathbcal{W}_1,...,\mathbcal{W}_C }{\operatorname{min}} \sum_{c=1}^{C} q_{c} \mathcal{L}_{\mathcal{D}_c}(\mathcal{h}_c(\mathbcal{W}_{c}, \mathbf{x}_c)) \label{eq4},
\end{equation}
where $q_c \geq 0$ and $\sum_c q_c=1$ denote the weights for each client when aggregating the local training updates. Based on \eqref{eq4}, the empirical objective function is
\begin{equation}
	\underset{\mathbcal{W}_1,...,\mathbcal{W}_C }{\operatorname{min}} N^{-1} \sum_{c=1}^{C} \sum_{j=1}^{n_c} \ell(\mathcal{h}_c(\mathbcal{W}_{c}, \mathbf{x}_{cj}), y_{cj}). \label{eq5}
\end{equation}

\subsection{Model estimation and prediction}
In this section, we discuss how to estimate the objective function \eqref{eq5}. We consider a two-layer fully connected ReLU activated neural network for illustration. Note that the optimization and prediction strategies provided here are also applicable to other DNNs. Assume the true decomposition of parameters is already known.
Suppose each of $C$ clients possesses $n$ training examples. Assume the FedSplit method is run for $T$ communication rounds. In each round, all clients are available with each of them training their local models for $G$ steps.
Denote $\mathbf{x} \in \mathbb{R}^d$ as the input. Since we consider a two-layer neural network as an example, let $\mathbf{W}_c$ denote the weight matrix of the first layer with $\mathbf{w}_r \in \mathbb{R}^d, r \in [m]$ representing the weight vector of the $r$th hidden node, and $m$ being the total number of hidden nodes.
Let $\mathbf{a} = (a_1, \ldots, a_m)^{\top}$ be the output weight vector. Define $\sigma(\cdot)$ to be the ReLU activation function, i.e., $\sigma(z) = z$ if $z \geq 0$ and $\sigma (z) = 0$ otherwise. Recall that we decompose the parameters of the first layer $\mathbf{W}_c$ into the shared ones $\mathbf{W}^s$ and client-specified ones $\mathbf{W}^p_c$, i.e., $\mathbf{W}_c=(\mathbf{W}^p_c, \mathbf{W}^s)$. To be more specific, assume we divide the hidden nodes of the first layer into
 $m_1$ personalized ones and $m_2$ shared ones, and denote $\mathbf{w}_{c,q}^p$ as the $q$-th ($q \in [m_1]$) personalized node
and $\mathbf{w}_{r}^s$ as the $r$-th ($r \in [m_2]$) shared node.
Then the neural network to be estimated is:
\begin{equation}
	h(\mathbf{W}_c, \mathbf{a}, \mathbf{x}) = \frac{1}{\sqrt{m_1}}\sum_{q=1}^{m_1}a_q\sum_{c=1}^{C}\sigma (\mathbf{w}_{c,q}^{p\top} \mathbf{x})\mathbb{I}\{\mathbf{x} \in D_c\}
	+ \frac{1}{\sqrt{m_2}}\sum_{r=1}^{m_2}a_r\sigma (\mathbf{w}_{r}^{s\top} \mathbf{x}). \label{theom-eq4}
\end{equation}

We can also understand model \eqref{theom-eq4} from representation learning. Denote the client-specific representation by $\psi^p_c: \mathbb{R}^d \rightarrow \mathbb{R}^{m_1}$, the common representation by $\psi^s: \mathbb{R}^d \rightarrow \mathbb{R}^{m_2}$, and the shared classifier by $f: \mathbb{R}^m \rightarrow \mathcal{Y}$. Then we have $\psi^p_c(\mathbf{x}) = \sigma\left(\mathbf{W}^p_c \mathbf{x}\right), \psi^s(\mathbf{x}) = \sigma\left(\mathbf{W}^s \mathbf{x}\right)$, and $f(\mathbf{v}) = \mathbf{a}^{\top}\mathbf{v}, \forall \mathbf{v} \in \mathbb{R}^{m}$ corresponding to \eqref{theom-eq4}.
The personalized model for the $c$-th client is the composition of the client’s shared head parameters and the common and unique representation: $h_c(\mathbf{x}) = \left(f \circ (\psi^s + \psi^p_c)\right)(\mathbf{x}) = \mathbf{a}^{\top}\left(\sigma\left(\mathbf{W}^p_c \mathbf{x}\right) + \sigma\left(\mathbf{W}^s \mathbf{x}\right)\right)$, where the symbol ``$+$'' means concatenation. Obviously, model \eqref{theom-eq4} focuses on incorporating the client-specific representation with the client-shared representation to customize the local model for heterogeneous clients.
This makes it remarkably distinct from a variety of personalized FL methods tailored for client-specific head learning or common representation learning \citep{liang2020think, collins2021exploiting, collins2022fedavg, oh2022fedbabu}.

In general, model \eqref{theom-eq4} can be optimized in a similar way with FedAvg but with different operations for client-shared parameters and client-specific parameters. Specifically, in each global communication round, the shared local updates of clients are aggregated by taking a simple average, while the personalized local updates are excluded from the averaging step. Below, we present the estimation procedure of \eqref{theom-eq4} in details.

Consider the quadratic loss function. Denote $S_c \subset [N]$ to be the index set of samples on client $c$ satisfying $S_i \cap S_j = \emptyset$ for $i \neq j$.
The local empirical objective function $l_c(\mathbf{W}_c)$ can be derived as follows:
\begin{equation*}
	l_c(\mathbf{W}_c) = \frac{1}{2n}\sum_{i \in S_c}(h(\mathbf{W}_c, \mathbf{a}, \mathbf{x}_i) - y_{i})^2.
\end{equation*}
The global empirical objective function then becomes:
\begin{equation}
	L(\mathbf{W}_1,\cdots, \mathbf{W}_C)=\frac{1}{C}\sum_{c=1}^{C}l_c(\mathbf{W}_c) \label{beq_1}.
\end{equation}
To minimize \eqref{beq_1}, we use the parallel gradient descent (GD) method. The local updating and global updating formulas are derived as follows.

\textbf{Local update.} Let $\mathbf{W}^p_{c,g}$/$\mathbf{W}^s_{c,g}$ represent the personalized/shared parameters of the $c$-th client in the $g$-th local update step.
 Then in each of $G$ local steps, $\mathbf{W}^p_{c,g}$ and $\mathbf{W}^s_{c,g}$ are updated by gradient descent, where
\begin{equation*}
	\mathbf{W}^p_{c,g+1} = \mathbf{W}^p_{c,g} - \eta_{l}\frac{\partial l_c(\mathbf{W}_c)}{\partial \mathbf{W}^p_{c,g}},~~
	\mathbf{W}^s_{c,g+1} = \mathbf{W}^s_{c,g} - \eta_{l}\frac{\partial l_c(\mathbf{W}_c)}{\partial \mathbf{W}^s_{c,g}},
\end{equation*}
where $\eta_l$ is the local learning rate. Let $\mathbf{w}^p_{c,g,q}$/$\mathbf{w}^s_{c,g,r}$ represent the parameters of the $q$-th/$r$-th personalized/shared neuron on the $c$-th client in the $g$-th local update step. The gradients are computed as follows,
\begin{equation*}
	\begin{aligned}
		\frac{\partial l_c(\mathbf{W}_c)}{\partial \mathbf{w}^p_{c,g,q}} &=\frac{1}{n\sqrt{m_1}}
		\sum_{i \in S_c}\left(h(\mathbf{W}_c, \mathbf{a}, \mathbf{x}_i) - y_{i}\right) a_{q} \mathbf{x}_{i}
		\mathbb{I}\left\{\mathbf{w}_{c,g,q}^{p\top} \mathbf{x}_{i} \geq 0\right\} ,\\
		\frac{\partial l_c(\mathbf{W}_c)}{\partial \mathbf{w}^s_{c,g,r}} &=\frac{1}{n\sqrt{m_2}}
		\sum_{i \in S_c}\left(h(\mathbf{W}_c, \mathbf{a}, \mathbf{x}_i) - y_{i}\right) a_{r} \mathbf{x}_{i}
		\mathbb{I}\left\{\mathbf{w}_{c,g,r}^{s\top} \mathbf{x}_{i} \geq 0\right\}.
	\end{aligned}
\end{equation*}

After $G$ local steps, the shared local updates are sent to the server for aggregation, and the personalized parameters are locally updated as:
\begin{equation*}
	\mathbf{W}_c^p(t+1) = \mathbf{W}_c^p(t) + \Delta \mathbf{W}_c^p(t).
\end{equation*}
Here $\mathbf{W}_c^p(t)$ denotes the personalized parameters of the $c$-th client in the $t$-th global update round, and $\mathbf{w}_{c,q}^p(t)$ is the corresponding parameters of the $q$-th personalized neuron. $\Delta \mathbf{W}_c^p(t)$ is the accumulated updates of the personalized parameters on the $c$-th client in the $t$-th round, with $\Delta \mathbf{w}^p_{c,q}(t)$ is the corresponding accumulated updates of the $q$-th personalized neuron which takes the following form:
\begin{equation*}
	\Delta \mathbf{w}^p_{c,q}(t) := a_q\frac{\eta_{l}}{\sqrt{m_1}} \sum_{g \in[G]}  \sum_{i \in S_c}\left(y_i-h_c^{(g)}(t)_i\right) \mathbf{x}_i \mathbb{I}\{\mathbf{w}^p_{c,g, q}(t)^{\top} \mathbf{x}_i \geq 0\}.
\end{equation*}
Here $\mathbf{h}^{(g)}_c(t) \in \mathbb{R}^M$ is the local model's outputs on the $c$-th client in the $g$-th local update step of the $t$-th communication round, and $h_c^{(g)}(t)_i$ is the $i$-th entry of $\mathbf{h}^{(g)}_c(t)$.

\textbf{Global update.} Denote $\Delta \mathbf{W}_c^s(t)$ as the accumulated updates of the shared parameters on the $c$-th client in the $t$-th round.
The shared local updates of all clients are aggregated as follows:
\begin{equation*}
	\Delta \mathbf{W}^s(t) = \sum_{c \in [C]} \frac{\Delta \mathbf{W}^s_c(t)}{C}.
\end{equation*}
Let $\eta_g$ denote the global learning rate, $\mathbf{W}^s(t)$ denote the shared parameters in the $t$-th global update round, and $\mathbf{w}_r^s(t)$ be the corresponding parameters of the $r$-th shared neuron. Then the shared parameters are globally updated as:
\begin{equation*}
	\mathbf{W}^s(t+1) = \mathbf{W}^s(t) + \eta_g \Delta \mathbf{W}^s(t),
\end{equation*}
where for $\forall c \in [C], r \in [m_2]$, we have
\begin{equation*}
	\Delta \mathbf{w}^s_r(t) :=\frac{a_r}{C} \sum_{c \in[C]} \sum_{g \in[G]} \frac{\eta_{l}}{\sqrt{m_2}} \sum_{i \in S_c}\left(y_i-h_c^{(g)}(t)_i\right) \mathbf{x}_i \mathbb{I}\{\mathbf{w}^s_{c,g, r}(t)^{\top} \mathbf{x}_i \geq 0\}.
\end{equation*}

Last, we discuss the prediction issue in federated learning. Practically, new clients often occur in the prediction stage. To accomplish the prediction task for new clients, we provide solutions for two typical scenarios. The first scenario is that the new clients can join the federated learning procedure. For these new clients, the personalized parameters $\mathbf{W}^p_c$ could be obtained by themselves through local training, and we just need to transfer the shared parameters $\mathbf{W}^s$ to the new clients to estimate the personalized models. This solution is also consistent with the practice in \cite{li2021fedbn}. We thus refer to this prediction strategy as \emph{LocalTrain}.
The second scenario is that the new clients cannot complete local training for some reasons, such as a lack of samples and insufficient computing power. In this case, assume the new clients have access to the personalized models of the existing clients \citep{marfoq2021federated, kamp2023federated}. Then the new clients can generate their predictions using the existing personalized models, and the outputs are ensemble together to gain the final prediction. We thus refer to this prediction strategy as \emph{Ensemble}.

\subsection{Theoretical properties}
In this section, we study the convergence and generalization property of model \eqref{theom-eq4} under a two-layer fully connected ReLU activated neural network model. To this end, we adopt the neural tangent kernel (NTK) technique \citep{jacot2018neural}, which is commonly used for convergence analysis of FL models \citep{huang2021fl, li2021fedbn, yue2022neural}. NTK is defined as the inner product space of the gradient of paired data points (also known as the Gram matrix). It describes the evolution of DNN during gradient descent training. Under the framework of NTK, the training of an infinite-width over-parameterized neural networks can be understood as a simple kernel gradient descent algorithm, and the kernel is fixed and only depends on the network structure and activation function. Therefore, the limit probability distribution to which the gradient descent converges can be seen as a stochastic process.
Here we follow \cite{huang2021fl} to study the theoretical properties of our proposed method. We first give the following assumption.
\begin{assumption}[Non-parallel data]
	\label{assum}
	For $i \in \{1, \ldots, N\}$, the data points $(\mathbf{x}_{i}, y_{i})$ satisfy $\|\mathbf{x}_{i}\|_2 \leq 1$ and $|y_i| \leq M$ for some constant $C_0$. For any $i \neq j$, we have $\mathbf{x}_{i} \nparallel \mathbf{x}_{j}$.
\end{assumption}
\noindent
The above assumption is essential to ensure the positive definiteness of the NTK/Gram matrix. This assumption is easy to satisfy because two inputs are hardly parallel in the context of horizontal federated learning. To facilitate the analysis of learning dynamics, we define the shared and personalized neural tangent kernels, which focus on cross-client and inner-client information respectively.
\begin{definition}[Neural tangent kernel]\label{define}
	The shared neural tangent kernel $\mathbf{H}^{s\infty} \in \mathbb{R}^{N \times N}$ and personalized tangent kernel $\mathbf{H}^{p\infty} \in \mathbb{R}^{N \times N}$ are given by
	\begin{equation*}
		\begin{aligned}
			\mathbf{H}^{s\infty}_{i,j}&:= \mathbb{E}_{\mathbf{w} \sim N(\mathbf{0}, \mathbf{I})}\left[\mathbf{x}_{i}^{\top} \mathbf{x}_{j} \mathbb{I}\left\{\mathbf{w}^{\top} \mathbf{x}_{i} \geq 0, \mathbf{w}^{\top} \mathbf{x}_{j} \geq 0\right\}\right],\\
			\mathbf{H}^{p\infty}_{i,j} &:= \mathbb{E}_{\mathbf{w} \sim N(\mathbf{0}, \mathbf{I})}\left[\mathbf{x}_{i}^{\top} \mathbf{x}_{j} \mathbb{I}\left\{\mathbf{w}^{\top} \mathbf{x}_{i} \geq 0, \mathbf{w}^{\top} \mathbf{x}_{j} \geq 0\right\} \mathbb{I}\{i \in S_c, j \in S_c, \forall c \in [C]\} \right].
		\end{aligned}
	\end{equation*}
\end{definition}
\noindent
Denote $\lambda^s$ and $\lambda^p$ as the smallest eigenvalues of $\mathbf{H}^{s\infty}$ and $\mathbf{H}^{p\infty}$,
respectively. According to \cite{du2018gradient}, we have $\lambda^s > 0$. Since $\mathbf{H}^{p\infty}$ is the diagonal matrix composed of the $C$ block matrices with size $n \times n$ on the diagonal of $\mathbf{H}^{s\infty}$, it is easily to infer that $\lambda^p \geq \lambda^s$.

We then study the convergence property of model \eqref{theom-eq4} trained using the multi-step parallel GD. In the traditional NTK theory, GD achieves zero training loss on shallow neural networks for regression tasks, and the convergence rate of infinite-width over-parameterized neural networks depends on the smallest eigenvalue of the induced kernel in the training evolution \citep{du2018gradient, jacot2018neural, arora2019fine, du2019gradient}. However, the analysis becomes much more complicated in our personalized federated learning. For one thing, we involve two types of parameters with different behaviors; for another, the movement of parameters are no longer directly determined by the local gradients.
 Following the common practice in NTK, we start from investigating the prediction trajectory of the neural network $h(\mathbf{W}_c, \mathbf{a}, \mathbf{x})$.
Denote $\mathbf{h}(t) \in \mathbb{R}^n$ to be the aggregated prediction vector of the personalized models after the $t$-th global round. Then the prediction errors are decomposed as
\begin{equation*}
	\begin{aligned}
		\|\mathbf{y}-  \mathbf{h}(t+1)\|^2_2
		& = \| \mathbf{y}-  \mathbf{h}(t) - \left( \mathbf{h}(t+1) - \mathbf{h}(t) \right) \|^2_2\\
		& = \| \mathbf{y}-  \mathbf{h}(t)\|^2_2 - 2 \left( \mathbf{y}-  \mathbf{h}(t)\right)^{\top}\left( \mathbf{h}(t+1) - \mathbf{h}(t) \right) + \| \mathbf{h}(t+1) - \mathbf{h}(t)\|^2_2.
	\end{aligned}
\end{equation*}
As show, the prediction errors in the $(t+1)$-th round are composed of those in the previous round (i.e., $\mathbf{y}- \mathbf{h}(t)$) and the difference between the predictions in adjacent rounds (i.e., $\mathbf{h}(t+1) - \mathbf{h}(t)$). To better understand the
learning dynamics, we separate the personalized/shared neurons into two sets, respectively. One set contains neurons with the activation pattern changing over time and the other set contains neurons with activation pattern remaining the same. Then we could analyze $\left( \mathbf{y}-  \mathbf{h}(t)\right)^{\top}\left( \mathbf{h}(t+1) - \mathbf{h}(t) \right)$ through investigating the personalized and shared Gram matrices, whose definitions are given as follows.

\begin{definition}[Gram matrix] The shared Gram matrix $\mathbf{H}(t,g,c)^s \in \mathbb{R}^{N \times N}$ and personalized Gram matrix $\mathbf{H}(t,g,c)^p \in \mathbb{R}^{N \times N}$ of the $c$-th client in the $g$-th local update step of the $t$-th communication round are given by
	\begin{equation*}
		\begin{aligned}
			\mathbf{H}(t,g,c)^s_{i,j} &= \frac{1}{m_2}\sum_{r=1}^{m_2}\mathbf{x}_i^{\top}\mathbf{x}_j \mathbb{I}\{\mathbf{w}^s_{c,g, r}(t)^{\top} \mathbf{x}_j \geq 0\}\mathbb{I}\{\mathbf{w}^s_{r}(t)^{\top} \mathbf{x}_i \geq 0\}, \\
			\mathbf{H}(t,g,c)^p_{i,j} &= \frac{1}{m_1} \sum_{q=1}^{m_1} \mathbf{x}_i^{\top}\mathbf{x}_j \mathbb{I}\{\mathbf{w}^p_{c,g, q}(t)^{\top} \mathbf{x}_j \geq 0\} \mathbb{I}\{\mathbf{w}^p_{c, q}(t)^{\top} \mathbf{x}_i \geq 0\}\mathbb{I}\left\{ i \in S_c\right\}.
		\end{aligned}
	\end{equation*}
\end{definition}

	Combining the $S_c$ columns of  $\mathbf{H}(t,g,c)^s$ for all $c \in [C]$, we get the shared Gram matrix $\mathbf{H}(t,g)^s \in \mathbb{R}^{N \times N}$ for the $g$-th local update step of the $t$-th communication round. Similarly, $\mathbf{H}(t,g)^p \in \mathbb{R}^{N \times N}$ is the personalized Gram matrix for the $g$-th local update step of the $t$-th communication round. Under appropriate assumptions as listed in Theorem~\ref{theorem2}, the evolving $\mathbf{H}(t,g)^p$/$\mathbf{H}(t,g)^s$ are close to their original Gram matrices $\mathbf{H}(t)^p$/$\mathbf{H}(t)^s$, the latter of which are close to $\mathbf{H}^{p\infty}$/$\mathbf{H}^{s\infty}$. On the other hand, by leveraging concentration properties at initialization, the difference between the prediction errors in local steps and those in the global step can be bounded. With these techniques, the term $\left( \mathbf{y}-  \mathbf{h}(t)\right)^{\top}\left( \mathbf{h}(t+1) - \mathbf{h}(t) \right)$ can be bounded by $\|\mathbf{y}-  \mathbf{h}(t)\|^2_2$. According to the classic NTK theory, the norm of the gradients can be controlled when the neural network is sufficiently wide. Thus the movement of the two kinds of parameters could be upper bounded. As a result, the term $ \| \mathbf{h}(t+1) - \mathbf{h}(t)\|^2_2$ is also controlled by the prediction errors. Consequently, the model \eqref{theom-eq4} could benefit from a linear learning rate, and the results are presented in the following theorem.

\setcounter{theorem}{0}
\renewcommand\thetheorem{\arabic{theorem}}
\begin{theorem}[Convergence Rate]
	\label{theorem2}
	Suppose Assumption \ref{assum} holds. 
	Let $\lambda^s = \lambda_{min}\left(\mathbf{H}^{s\infty}\right)$, $\gamma^s = \lambda_{max}\left(\mathbf{H}^{s\infty}\right)/\lambda_{min}\left(\mathbf{H}^{s\infty}\right)$,
	$\lambda^p = \lambda_{min}\left(\mathbf{H}^{p\infty}\right)$, $\gamma^p = \lambda_{max}\left(\mathbf{H}^{p\infty}\right)/\lambda_{min}\left(\mathbf{H}^{p\infty}\right)$, $\rho_1 = \eta_g\eta_lG/C$, $\rho_2 = \eta_lG$. Fix $m_1=\Omega\left(((\lambda^p)^{-4}C^{-2}) N^4 \log (N / \delta)\right)$ and $m_2=\Omega\left((\lambda^s)^{-4} N^4 \log (N / \delta)\right)$. Furthermore, we i.i.d. initialize $\mathbf{w}^p_{c,q}$ and $\mathbf{w}^s_r \sim \text{MN}(\mathbf{0}, \mathbf{I})$, and sample $a_q, a_r$ from $\{-1, 1\}$ uniformly at random for
	$c\in [C], q\in [m_1], r \in [m_2]$, then with probability at least $1 - \delta$, training network \eqref{theom-eq4} using the GD algorithm with global step size $\eta_g = O(1)$ and local step size $\eta_l =O\left(\frac{\lambda^s + \lambda^p}{\gamma^s\gamma^pGN^2}\right)$ converges linearly as
	\begin{equation*}
		\|\mathbf{h}(t)-\mathbf{y}\|_{2}^{2} \leq\left(1- \frac{1}{2}\left(\rho_1\lambda^s+\rho_2\lambda^p\right) \right)^{t}\|\mathbf{h}(0)-\mathbf{y}\|_{2}^{2}.
	\end{equation*}
\end{theorem}

The detailed proof of Theorem~\ref{theorem2} can be found in Supplementary Materials B.
This theorem shows that the convergence guarantees of model \eqref{theom-eq4} are governed by the spectral properties of both the shared NTK and personalized NTK. According to \cite{huang2021fl}, the convergence rate of the two-layer
fully connected ReLU activated neural network trained with FedAvg is $O\left(1 - \rho^{\prime}\lambda^s/2 \right)$, where $\rho^{\prime} = \eta_g\eta_{l}^{\prime}G/C$ and $\eta_{l}^{\prime} = O\left(\lambda^s/ (\gamma^s G N^2)\right)$. We have $\rho^{\prime} \approx \rho_1$ when the total sample size $N$ is large. Notably, the convergence rate of FedSplit is $O\left(1- \rho_1 \lambda^s/2 - \rho_2\lambda^p/2\right)$, where the additional term $\rho_2\lambda^p/2$ is introduced due to layer partition.
Therefore, FedSplit enjoys faster convergence rate than FedAvg.
	It implies that, in our FedSplit method, the update in the customized model is not only determined by the averaged gradient directions, but also by the direct gradient direction from heterogeneous clients, so that the parameters are optimized towards the most favorable path for a specific client.
	In addition, our proposed FedSplit model enjoys linear convergence rate under the neural network setting, without assumptions on the convexity of objective functions or
	distribution of data. On the contrary, existing methods have to rely on the assumptions of convexity and smoothness for the objective functions to develop their convergence guarantees. These assumptions are not realistic for non-linear neural networks. For example, \cite{t2020personalized} demonstrated that, their method pFedMe could obtain a sublinear speed of $\mathcal{O}\left(1 /(T G C)^{2 / 3}\right)$ for $L$-smoothed nonconvex objectives (Assumption 1), with bounded variance of gradients (Assumption 2) and bounded gradient diversity (Assumption 3) between the global model and local models. Except the common assumptions on smoothness and gradients, \cite{fallah2020personalized} also required the loss functions be bounded and twice continuously differentiable, and its convergence speed was $\mathcal{O}\left(1 /(T)^{3/2}\right)$.

Last, we study the upper bound for the generalization error of the FedSplit method. To achieve this goal, additional conditions on the training data distribution and test data distribution are required. The details are summarized in Theorem~\ref{theorem3}.

\begin{theorem}[Generalization Bounds]\setcounter{part}{2}
	\label{theorem3}
	Fix failure probability $\delta \in(0,1)$. Suppose the training data $S=\left\{\left(x_i, y_i\right)\right\}_{i=1}^N$ in the FL system are i.i.d. samples from $\otimes^C_{c=1}\mathcal{D}_c$, such that with probability at least $1 - \delta/3$, we have $\lambda_{min}\left(\mathbf{H}^{s\infty}+ \mathbf{H}^{p\infty}\right) \geq \lambda >0$. Let $\lambda^s = \lambda_{min}\left(\mathbf{H}^{s\infty}\right)$, $\lambda^p = \lambda_{min}\left(\mathbf{H}^{p\infty}\right)$, $\rho_1 = \eta_g\eta_lG/C$, and $\rho_2 = \eta_lG$.
	Fix $\alpha=O(\lambda \operatorname{poly}(\log N, \log (1 / \delta)) / N)$,
	$m_1= \Omega\left(\alpha^{-2}\left(\operatorname{poly}\left(N,\log (1 / \delta), (\lambda^p)^{-1}, \lambda^{-1}\right)\right)\right)$, and $m_2 = \Omega\left(\alpha^{-2}\left(\operatorname{poly}\left(N,\log (1 / \delta), (\lambda^s)^{-1}, \lambda^{-1}\right)\right)\right)$.
	Let the
	neural network \eqref{theom-eq4} be initialized with $\mathbf{w}^p_{c,q}$ and $\mathbf{w}^s_r$, which are i.i.d. sampled from $\text{MN}(\mathbf{0}, \alpha^2\mathbf{I})$. Let $a_q, a_r$ be sampled from $\{-1, 1\}$ uniformly for
	$c\in [C], q\in [m_1], r \in [m_2]$. Assume we train model \eqref{theom-eq4} in the FL framework for $T \geq \Omega\left(\left(\rho_1\lambda^s+ \rho_2\lambda^p\right)^{-1} \operatorname{poly}(\log (N / \delta))\right)$ iterations.
	Consider the loss function $\ell: \mathbb{R} \times \mathbb{R} \rightarrow[0,1]$ that is 1-Lipschitz in its first argument. Then with probability at least $1-\delta$ over the random initialization on $\mathbf{W}^s(0) \in \mathbb{R}^{d \times m_2}, \mathbf{W}^p(0) \in \mathbb{R}^{d \times Cm_1}$ and the training samples, the population loss $\mathcal{L}_{\otimes^C_{c=1}\mathcal{D}_c}(h):=\mathbb{E}_{(\mathbf{x}, y) \sim \otimes^C_{c=1}\mathcal{D}_c}[\ell(h(\mathbf{W}, \mathbf{a}, \mathbf{x}), y)]$ is upper bounded by
	$$
	\mathcal{L}_{\otimes^C_{c=1}\mathcal{D}_c}(h) \leq 2\sqrt{(C/n) \mathbf{y}^{\top}\left(\mathbf{H}^{s\infty}+\mathbf{H}^{p\infty}\right)^{-1} \mathbf{y}}+O(\sqrt{\log (N /(\lambda \delta)) /(2 N)}).
	$$
\end{theorem}

The detailed proof of Theorem~\ref{theorem3} can be found in Supplementary Materials C.
	The intuitions behind Theorem~\ref{theorem3} are that, if the proposed model \eqref{theom-eq4} satisfies the listed conditions, then its generalization bound depends on a data-dependent complexity measure (i.e., the first term on the right side of the inequality) and is unrelated with the network itself. The complexity measure is determined by both the cross-client information and inner-client information, as well as the number of clients $C$ and sample size on each client $n$. Given the total training sample size $N$ in the FL system, increasing the number of clients $C$ or decreasing the number of local samples $n$ on each client would lead to larger generalization error. Intuitively, a large client number on a fixed amount of data means it is more difficult to develop personalized models that could generalize well for all clients.

\section{Factor-Assisted Decomposition}
\label{fad}
\subsection{Decomposition with factor analysis}
A key problem to optimize the objective function \eqref{eq5} is to decompose the whole parameter set into client-shared ones and client-specific ones. To achieve this goal, we apply the \emph{factor analysis} technique \citep{harman1976modern}. Factor analysis is a classic statistical model and is widely used in fields such as sociology, political science, and economics \citep{kirkegaard2016inequality,abduljaber2020dimension,song2021evaluation}. It assumes there exists inter-dependence among a large set of variables, which can be characterized by several ``common latent factors". In addition, these common latent factors often have different influencing levels to the variables. Then variables influenced much by the common latent factors can be regarded as a shared group. In our problem, we treat the hidden elements in each layer as variables. Then factor analysis is conducted to find the common latent factors among the hidden elements. The hidden elements scoring high on the common factors would be regarded as client-shared ones, whereas the others would be regarded as client-specific ones. We then discuss the factor-assisted decomposition procedure in detail.


For the $l$-th ($l \in \Lambda$) convolutional layer, assume it has $d_l$ kernels with a fixed size $s_1 \times s_2$. Then the depth of the feature map in the $(l+1)$th layer is $d_{l}$. Let $\mathcal{K}_l$ denote the set of $d_l$ kernels. Assume there exists inter-dependence among $\mathcal{K}_l$, which can be characterized by $G_l$ latent factors $\mathbf{F}_l = (F_1, \cdots, F_{G_l})^{\top}$ with $G_l<d_l$. Then the factor model is
\begin{equation}
	\label{FA}
	\mathcal{K}_l = \mathbf{A}_l\mathbf{F}_l + \mathbf{\epsilon}_l,
\end{equation}
\noindent
where $\mathbf{A}_l=(a_{ij})\in \mathbb{R}^{d_l \times G_l}$ denotes the loading matrix, and $\mathbf{\epsilon}_l = (\epsilon_1, \cdots, \epsilon_{d_l})^{\top}$ represents the information that cannot be explained by $G_l$ factors.

The factor analysis is conducted on the server side. Before that, the received intermediate updates tensors (weights or gradients) from clients should be reshaped and combined to form an input matrix for the factor model, which are shown in Figure~\ref{fig2}.
Assume the server selects $K \leq N$ active clients for the current communication round, denoted by $\mathcal{S}_t$. Then for each client, the intermediate updates tensors with size $d_{l-1} \times s_1 \times s_2$ of each of the $d_l$ kernel is flattened to form a weight vector {$\mathbf{z}_{clj} \in \mathbb{R}^{d_{l-1}s_{1}s_{2}}$ with $c \in \mathcal{S}_t$ and $j \in [d_l]$. Second, the flattened updates vectors of $K$ active clients are then concatenated client-wise to form a longer vector $\mathbf{\chi}_{lj}=\left(\mathbf{z}_{1 lj}, \mathbf{z}_{2l j}, \cdots, \mathbf{z}_{Kl j}\right)^{\top}\in\mathbb{R}^{d_{l-1}s_{1}s_{2}K} $. Third, the $d_l$ concatenated vectors are combined to form a matrix, which is denoted by $\mathbf{Z}_l=\left(\mathbf{\chi}_{l1}, \mathbf{\chi}_{l2}, \cdots, \mathbf{\chi}_{ld_{l}}\right) \in \mathbb{R}^{(d_{l-1}s_{1}s_{2}K) \times d_l}$. After these steps, $\mathbf{Z}_l$ is the input for factor analysis on the $l$-th layer.
	
	\begin{figure}[ht]
		\centering
		\includegraphics[width=0.8\textwidth]{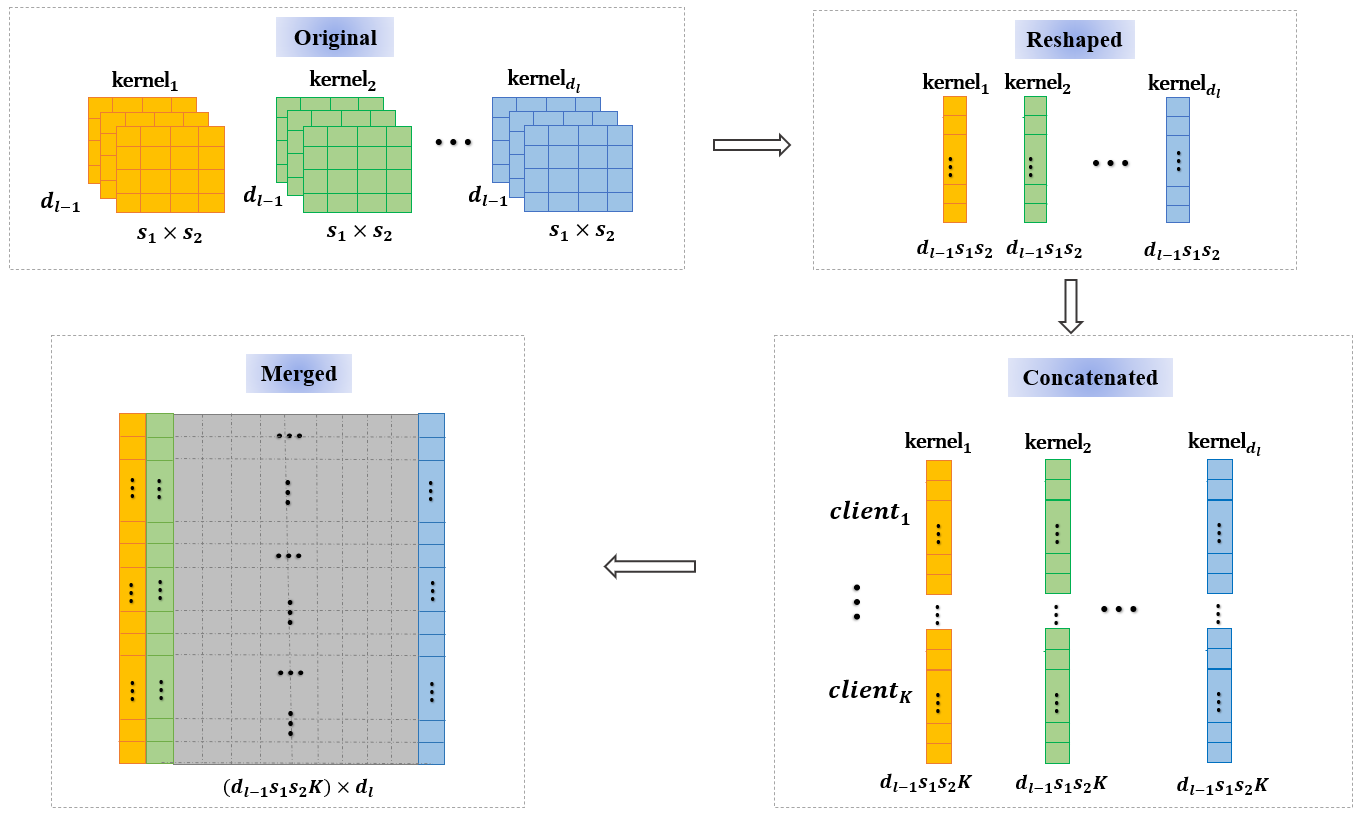}
		\caption{Data preparation for factor analysis.}
		\label{fig2}
	\end{figure}

	We follow the common practice to estimate the loading matrix $\mathbf{A}_l$, as well as choosing the appropriate number of latent factors $G_l$ \citep{harman1976modern}.
	First, we normalize each column of the input matrix and still denote
	$\mathbf{Z}_l$ as the normalized one. Then we compute the correlation matrix of $\mathcal{K}_l$ as $R_l = \mathbf{Z}_l^{\top}\mathbf{Z}_l$. Assume $\gamma_{1} \geq \gamma_2 \geq \cdots \geq \gamma_{d_l} \geq 0$ are the eigenvalues of $R_l$ and $u_{j} (j=1,\cdots,d_l)$ is the corresponding eigenvector. To determine the appropriate number of latent factors $G_l$, we compute a cumulative variability contribution rate as $r_{lm}=\sum_{j=1}^{m} \gamma_{j} / \sum_{j=1}^{d_{l}} \gamma_{j}$ for $1 \leqslant m \leqslant d_l$. Then define $G_l = \min\{m: r_{lm} \geqslant \kappa_l\}$, where $\kappa_l$ is a pre-specified threshold. Denote $\Sigma_{\mathbf{\epsilon}_l}$ as the covariance matrix of special factors.
	According to classic theory of factor model, we have $R_l = \mathbf{A}_l\mathbf{A}_l^{\top}+\Sigma_{\mathbf{\epsilon}_l}$. Let $R^*_l = R_l - \Sigma_{\mathbf{\epsilon}_l}$. Assume the top positive $G_l$ eigenvalues of $R^*_l$ are $\gamma^*_{1}, \gamma^*_2 ,\cdots,\gamma^*_{G_l}$, and $\mathbf{u}^*_{j} (j=1,\cdots,G_l)$ is the corresponding eigenvector.
	Then $\mathbf{A}_l$ can be estimated as
	\begin{equation}\label{estimate_A}
		\widehat{\mathbf{A}}_l = \left(\sqrt{\gamma^*_{1}}\mathbf{u}^*_{1}, \cdots, \sqrt{\gamma^*_{G_l}}\mathbf{u}^*_{G_l} \right)^{\top}.
	\end{equation}
	Since $\Sigma_{\mathbf{\epsilon}_l}$ is unknown, we execute the estimation process literately. First we obtain $\widehat{\mathbf{A}}_l$ using \eqref{estimate_A}. Then we compute $\widehat{\Sigma}_{\mathbf{\epsilon}_l}= R_l - \widehat{\mathbf{A}}_l\widehat{\mathbf{A}}_l^{\top}$. The process is iterated until the difference of $\widehat{\Sigma}_{\mathbf{\epsilon}_l}$ between consecutive two iterations is very small. Here we could initialize $\widehat{\mathbf{A}}_l$ as $\left(\sqrt{\gamma_{1}}\mathbf{u}_{1}, \cdots, \sqrt{\gamma_{G_l}}\mathbf{u}_{G_l} \right)^{\top}$.

	Based on the estimation results of $\mathbf{A}_l$, we split the hidden elements in the $l$-th layer as follows. According to the property of factor analysis, we have $\sum_{m=1}^{d_l}a_{jm}^2=1$ for the $j$th kernel. Then we define a new measure $\nu_{lj}=\sum_{m=1}^{G_l}a_{jm}^2$ to indicate the variation proportion of the $j$th kernel that can be explained by $G_l$ common factors.
	Then a higher $\nu_{lj}$ indicates the $j$th kernel could be more explained by the common factors, and thus more likely to be shared by all clients. This idea can be understood from another perspective. Assume there exists a kernel $j'$ with all $a_{j'm}=0$. Since this kernel has zero loadings on all common factors, it is immune to general knowledge and behaves as an absolutely client-specific one. Then we can rewrite $\nu_{lj}=\sum_{m=1}^{G_l}(a_{jm}-0)^2$, which is the Euclidean distance between the $j$th kernel and the $j'$th kernel. Thus the $j$th kernel with smaller $\nu_{lj}$ suggests it is more similar to the $j'$th kernel and behaves more client-specific.
	To split the kernels, let $\zeta_{lj} \in\{0,1\}$ be an indicator variable, where $\zeta_{lj}=1$ denotes the $j$th kernel in the $l$-th layer to be client-shared and $\zeta_{lj}=0$ otherwise. By using a threshold $\tau_l$, define $\zeta_{l j}=1$ if $v_{l j} \geq \tau_l$ and $\zeta_{l j}=0$ if $v_{l j} < \tau_l$.
	In this way, each element in the hidden layer can be categorized into the client-shared group $\boldsymbol{I}^{s}_{l} = \{j: \zeta_{lj} = 1, 1\leq j \leq d_l\}$ and client-specific group $\mathbf{I}^{p}_{l} = \{j: \zeta_{lj} = 0, 1\leq j \leq d_l\}$. All convolution layers and the dense layer can be split sequentially.

With the factor analysis, we can split the whole parameter set into the client-shared ones and client-specific ones. Then the objective function \eqref{eq5} of FedSplit can be optimized. We refer to the FedSplit method with factor analysis used for decomposition as \emph{FedFac}, which is a practically implemented version of FedSplit. The framework of FedFac is summarized in Figure~\ref{fig3}.
	
	\begin{figure}[h]
		\centering
		\includegraphics[width=0.95\textwidth]{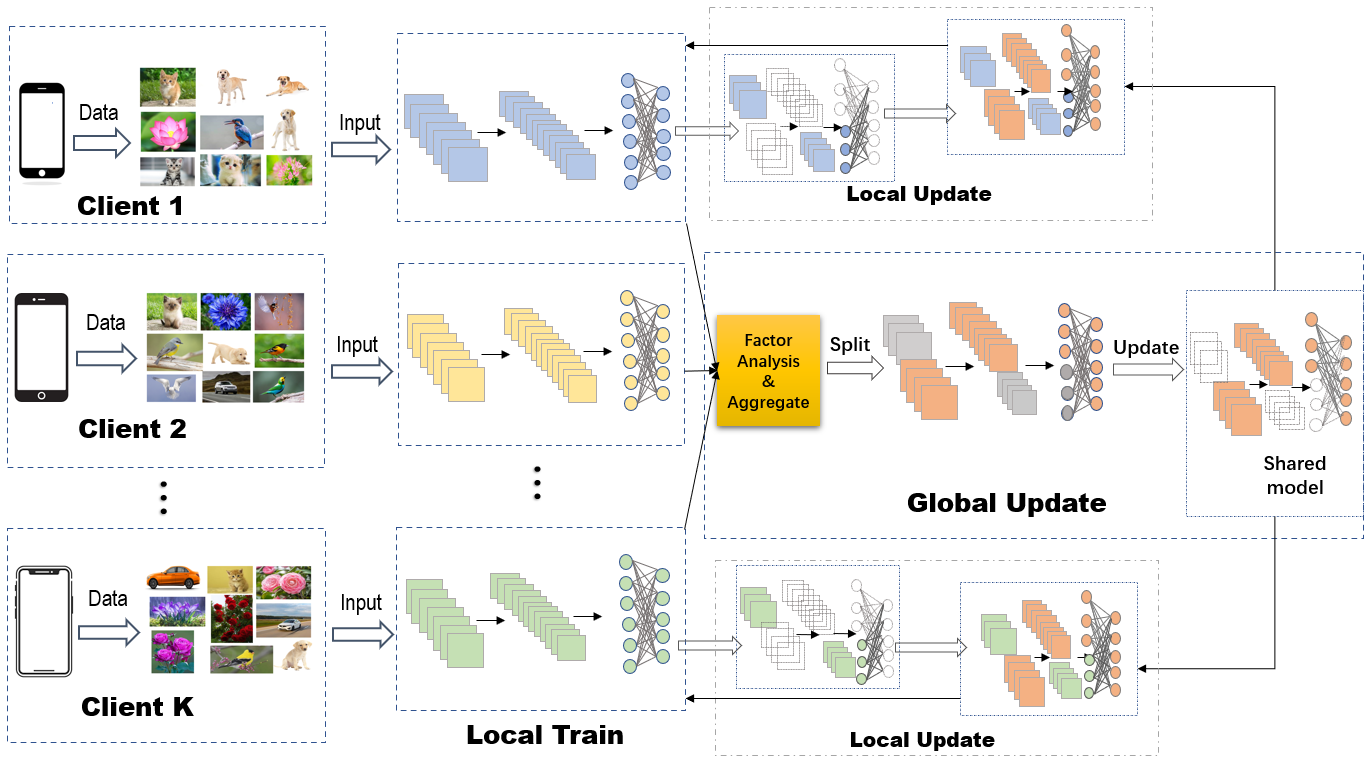}
		\caption{The overall framework of FedFac.}
		\label{fig3}
	\end{figure}	

	\subsection{Static and dynamic algorithms}
	\label{sec1}
To implement factor analysis to decompose the hidden elements into client-shared group and client-specific group, we develop two algorithms, i.e., the static FedFac algorithm and the dynamic FedFac algorithm. In short, the static FedFac algorithm only conducts factor analysis once and then fixes the decomposition. Specifically, assume a certain number of clients can first train local models based on their private datasets. Then the server gathers the local parameters of specific layers and then conducts factor analysis on the aggregated weight matrix. As a result, we could discriminate the shared hidden elements (neurons or channels) from personalized ones based on common and heterogeneous knowledge. Then the decomposition of parameters is fixed during the whole FL procedure. We call this algorithm the static FedFac (Alg.~\ref{alg:one}) since factor analysis is only conducted once in the initialization step. This static algorithm is convenient to execute. In addition, since the decomposition of hidden elements is fixed during the training procedure, our theoretical analysis of FedSplit is applicable.

	\begin{algorithm}[h]
		\footnotesize
		\caption{The Static FedFac Algorithm}\label{alg:one}
		\KwData{$T, \eta_g, K, \Lambda, \tau_l, \kappa_l, (l \in \Lambda)$}
		\KwResult{$\mathbcal{W}_c = \left(\mathbcal{W}^{s},\mathbcal{W}^{p}_c\right)$ for each $c \in [C]$}
		\textbf{Initialization:} The server collects the parameters of $l$-th layer for $l \in \Lambda$ from each local model, conducts factor analysis on the aggregated weight matrix to decide parameters partition $\boldsymbol{I}^{s}_{l}$ and $\boldsymbol{I}^{p}_{l}$ for each $l \in \Lambda$ \;
		Let $\mathbcal{W}^{s} = \{\mathbcal{w}_a, \{\mathbcal{w}^s_{lj}\}\}, a \notin \Lambda, l \in \Lambda, j \in \boldsymbol{I}^{s}_{l}$; and
		$\mathbcal{W}^{p} = \{\mathbcal{w}^p_{lj}\}, l \in \Lambda, j \in \boldsymbol{I}^{p}_{l}$ \;
		Initialize $\mathbcal{W}_{c} = \{\mathbcal{W}^{s}(0), \mathbcal{W}^{p}_{c}(0)\}, c\in[C]$ \;
		\For {each round $t = 1,2,\cdots,T$}{
			$\mathcal{S}_t \leftarrow$ (random set of $K$ clients)\;
			\bf broadcasts $\mathbcal{W}^{s}(t)$ to clients in $\mathcal{S}_t$\;
			\For{each client $k \in \mathcal{S}_t$}{
				$\left( \Delta \mathbcal{W}^s_{k}(t), \Delta \mathbcal{W}^p_{k}(t)\right) \leftarrow$ ClientUpdate $\left(k, \mathbcal{W}^{s}(t), \mathbcal{W}^{p}_{k}(t) \right)$ \;
				$\mathbcal{W}^p_{k}(t+1) = \mathbcal{W}^p_{k}(t) + \Delta \mathbcal{W}^p_{k}(t)$
			}
			$\Delta \mathbcal{W}^s(t) = \sum_{k \in \mathcal{S}_t}\frac{n_k \Delta \mathbcal{W}^s_{k}(t)}{\sum_{k \in \mathcal{S}_t} n_k} $ \;
			$\mathbcal{W}^s(t+1) = \mathbcal{W}^s(t) + \eta_g \Delta \mathbcal{W}^s(t)$
		}
		\textbf{\normalsize{Client Update}} \\
		\KwData{client index $c$, local epoches $G$, learning rate $\eta_l$ and batchsize $B$}
		\KwResult{$\mathbcal{W}_c$}
		download $\mathbcal{W}^{s}$ from server\;
		$\mathbcal{W}_{c}=\{\mathbcal{W}^{s}, \mathbcal{W}^{p}_{c}\}$\;
		$\mathcal{B} \leftarrow\left(\right.$ split $\mathcal{D}_{c}$ into batches of size $\left.B\right)$\;
		\For{each local epoch $g$ from $1$ to $G$}{
			\For{batch $b \in \mathcal{B}$}{
				$\mathbcal{W}_c \leftarrow \mathbcal{W}_c-\eta \nabla \ell(\mathbcal{W}_c; b)$
			}
		}
	\end{algorithm}
	
However, by using the static method, we have no chance to adjust the initial decomposition if it is far away from optimal. To address this issue, we further develop a dynamic version of FedFac (Alg.~\ref{alg:two}). The algorithm of dynamic FedFac conducts factor analysis in each communication round so as to modify decomposition based on the newest common knowledge the FL system has learned. Specifically, a subset of clients are selected at each communication round to perform local updates based on their personalized models. Then the updated intermediate parameters are sent to the server, where the interested parameters are reshaped and factor analysis is performed to decide parameter decomposition. After that, the shared parameters are averaged to form a global update, whereas the personalized parameters are updated by each client. Compared with the static method, the dynamic method is more flexible.
		
	\begin{algorithm}[h]
		\footnotesize
		\caption{The Dynamic FedFac Algorithm}\label{alg:two}
		\KwData{$T, \eta_g,K, \Lambda, \tau_l, \kappa_l, (l \in \Lambda)$}
		\KwResult{$\mathbcal{W}_c = \left(\mathbcal{W}^{s},\mathbcal{W}^{p}_c\right)$ for each $c \in [N]$}
		\textbf{Initialization:} The server collects the parameters of $l$-th layer for $l \in \Lambda$ from each local model, conducts factor analysis on the aggregated weight matrix to decide parameters partition $\boldsymbol{I}^{s}_{l}$ and $\boldsymbol{I}^{p}_{l}$ for each $l \in \Lambda$ \;
		Let $\mathbcal{W}^{s} = \{\mathbcal{w}_a, \{\mathbcal{w}^s_{lj}\}\}, a \notin \Lambda, l \in \Lambda, j \in \boldsymbol{I}^{s}_{l}$; and
		$\mathbcal{W}^{p} = \{\mathbcal{w}^p_{lj}\}, l \in \Lambda, j \in \boldsymbol{I}^{p}_{l}$ \;
		Initialize $\mathbcal{W}_{c} = \{\mathbcal{W}^{s}(0), \mathbcal{W}^{p}_{c}(0)\}, c\in[N]$ \;
		\For {each round $t = 1,2,\cdots,T$}{
			$\mathcal{S}_t \leftarrow$ (random set of $K$ clients)\;
			\bf broadcasts $\mathbcal{W}^{s}(t)$ to clients in $\mathcal{S}_t$\;
			\For{each client $k \in \mathcal{S}_t$}{
				$\Delta \mathbcal{W}_{k}(t) \leftarrow$ ClientUpdate $\left(k, \mathbcal{W}^{s}(t), \mathbcal{W}^{p}_{k}(t) \right)$ \;
			}
			\For{each layer $l \in \Lambda$}{
				conducts factor analysis to decide parameters partition $\boldsymbol{I}^{s}_{l}(t)$ and $\boldsymbol{I}^{p}_{l}(t)$
			}
			$\Delta \mathbcal{W}^{s}_k(t) = \{\Delta \mathbcal{w}_{ka}, \{\Delta \mathbcal{w}^s_{klj}\}\}, \text{for}~ k \in \mathcal{S}_t, a \notin \Lambda, l \in \Lambda, j \in \boldsymbol{I}^{s}_{l}(t)$ \;
			$\Delta \mathbcal{W}^s(t) = \sum_{k \in \mathcal{S}_t}\frac{n_k \Delta \mathbcal{W}^s_{k}(t)}{\sum_{k \in \mathcal{S}_t} n_k} $ \;
			$\mathbcal{W}^s(t+1) = \mathbcal{W}^s(t) + \eta_g \Delta \mathbcal{W}^s(t)$\;
			$\mathbcal{W}^{p}_k(t+1) = \{\mathbcal{w}^p_{klj}(t) + \Delta \mathbcal{w}^p_{klj}(t)\}, \text{for}~ k \in \mathcal{S}_t, l \in \Lambda, j \in \boldsymbol{I}^{p}_{l}(t)$ \;
		}
		\textbf{\normalsize{Client Update}} \\
		\KwData{client index $c$, local epoches $G$, learning rate $\eta_l$ and batchsize $B$}
		\KwResult{$\mathbcal{W}_c$}
		download $\mathbcal{W}^{s}$ from server\;
		$\mathbcal{W}_{c}=\{\mathbcal{W}^{s}, \mathbcal{W}^{p}_{c}\}$\;
		$\mathcal{B} \leftarrow\left(\right.$ split $\mathcal{D}_{c}$ into batches of size $\left.B\right)$\;
		\For{each local epoch $g$ from $1$ to $G$}{
			\For{batch $b \in \mathcal{B}$}{
				$\mathbcal{W}_c \leftarrow \mathbcal{W}_c-\eta \nabla \ell(\mathbcal{W}_c; b)$
			}
		}
	\end{algorithm}
	
\subsection{Effectiveness of using factor analysis}\label{Simulation}
In this section, we investigate whether using factor analysis can recover the true underlying decomposition. To study this problem, we design a simulation study to compare the performance of different decomposition methods. Below, we first present the simulation setup and then present the corresponding results.

\subsubsection{Simulation Setup}

We first present the data generation process, which can generate heterogeneous data suitable for FedFac. For $c\in[C]$, denote $\mathbf{x}_c = (\mathbf{x}_c^p, \mathbf{x}_c^s) \in \mathbb{R}^d$, where $\mathbf{x}_c^p \in \mathbb{R}^{d_1}$ represents the personalized covariates and $\mathbf{x}_c^s \in \mathbb{R}^{d_2}$ represents the shared covariates. Let $\alpha = d_2/d$ indicate the shared proportion of $\mathbf{x}_c$. Assume $\mathbf{x}_c^s \sim \text{MN}(\mathbf{0}, \mathbf{I})$ and $\mathbf{x}_c^p \sim \text{MN}(\boldsymbol{\mu}_c, \mathbf{\Sigma})$, where $\text{MN}(\cdot)$ represents the multi-normal distribution and $\boldsymbol{\mu}_c \sim \text{MN}(\mathbf{0}, \mathbf{I}), \mathbf{\Sigma} = (0.5^{|i-j|})_{d_1 \times d_1}$. Let $\mathbf{W}^s = \left(\left(\mathbf{w}^{sp}\right)^{\prime}, \left(\mathbf{w}^{ss}\right)^{\prime}\right)^{\prime} \in \mathbb{R}^{d \times m_2}$ denote the shared parameters, and $\mathbf{W}^p_c = \left(\left(\mathbf{w}^{pp}_c\right)^{\prime}, \left(\mathbf{w}^{ps}_c\right)^{\prime}\right)^{\prime} \in \mathbb{R}^{d \times m_1}$ denote the personalized parameters for the $c$-th client. Denote $p = m_2/m, m = m_1 + m_2$. Then $p$ represents the shared proportion of model parameters. Denote $\mathbf{a} = (a_1, \cdots, a_m)^{\top} \in \mathbb{R}^{m}$.
Then the true personalized model of each client is assumed as follows:
\begin{equation}
	h^{*}_c(\mathbf{x}_c) = \sum_{q=1}^{m_1}a_q\sigma \left(\left(\mathbf{w}_{cq}^{pp}\right)^{\prime} \mathbf{x}_c^p + \left(\mathbf{w}_{cq}^{ps}\right)^{\prime} \mathbf{x}_c^s\right) + \sum_{r=1 + m_1}^{m_1 + m_2}a_r\sigma \left(\left(\mathbf{w}_{r}^{sp}\right)^{\prime} \mathbf{x}_c^p + \left(\mathbf{w}_{r}^{ss}\right)^{\prime} \mathbf{x}_c^s\right). \label{De}
\end{equation}

We fix $\mathbf{W}^s$ and $\mathbf{a}$ for all clients, with $\mathbf{w}^{sp}_r \sim U(-0.1, 0.1)$, $\mathbf{w}^{ss}_r \sim U(-1, 1)$ for $r \in [m_1+1:m_1+m_2]$, and $\mathbf{a} \sim \text{MN}(\mathbf{0}, \mathbf{I})$. The personalized parameters are set client-specific with $\mathbf{w}^{pp}_{cq} \sim \text{MN}(\boldsymbol{\mu}_c, \mathbf{I})$ and $\mathbf{w}^{ps}_{cq} \sim U(-0.1, 0.1)$ for $q \in [m_1]$, for the $c$-th client.
Assume $y_c^*$ is generated as $y_c^* = h^{*}_c(\mathbf{x}_c) + \epsilon$ where $\epsilon \sim N(0, \mathcal{c}^2)$. Then to generate a binary  variable $y_c$, we consider the sigmoid function $f(x)$. Specifically, we set $y_c=1$ if $f(y_c^*) \in (0.5, 1]$; otherwise $y_c=0$.

In the simulation study, we set $C = 100, d = 100$, and $m = 200$. Then the MLP with one hidden layer of $m$ neurons is used to fit the generated data.
For comparison purpose, we consider five decomposition methods. They are, respectively: (1) the true decomposition, (2) the oracle decomposition, where factor analysis is conducted on the true parameters, (3) the estimated decomposition using static FedFac, (4) the estimated decomposition using dynamic FedFac, and (5) the random decomposition. In the true decomposition method, we know the true indices of personalized and shared hidden elements, which are stated in \eqref{De}.
In the oracle decomposition method, we conduct decomposition using factor analysis based on the
true parameters of all clients, i.e., $\mathbf{W}_c$ = $\left(\mathbf{W}^p_c, \mathbf{W}^s\right) \in \mathbb{R}^{d \times m}, c \in [C]$, and then execute  FL training under the proposed framework.
In the estimated decomposition methods (static FedFac or dynamic FedFac), we neither know the true decomposition nor the true parameters. It is the same case we encountered in practice. In the random decomposition, we split shared/personalized neurons by random guess. Note that in the true and oracle methods, the structure of DNN stays the same throughout the FL training. Therefore, we only average the parameters of the shared neurons in each communication round.

\subsubsection{Simulation results}

We consider different experimental settings by changing the values of $p$ (the shared proportion of model parameters) and $\alpha$ (the shared proportion of covariates). Figure~\ref{simu1} plots the testing accuracies using different decomposition methods. As shown, the first four decomposition methods can generate nearly identical results, while the random decomposition is much inferior. This finding suggests that, the estimated decomposition methods (both static FedFac and dynamic FedFac) used in practice can well approximate the true decomposition with high accuracy.

\begin{figure}[h]
	\centering
	\includegraphics[width=0.9\textwidth]{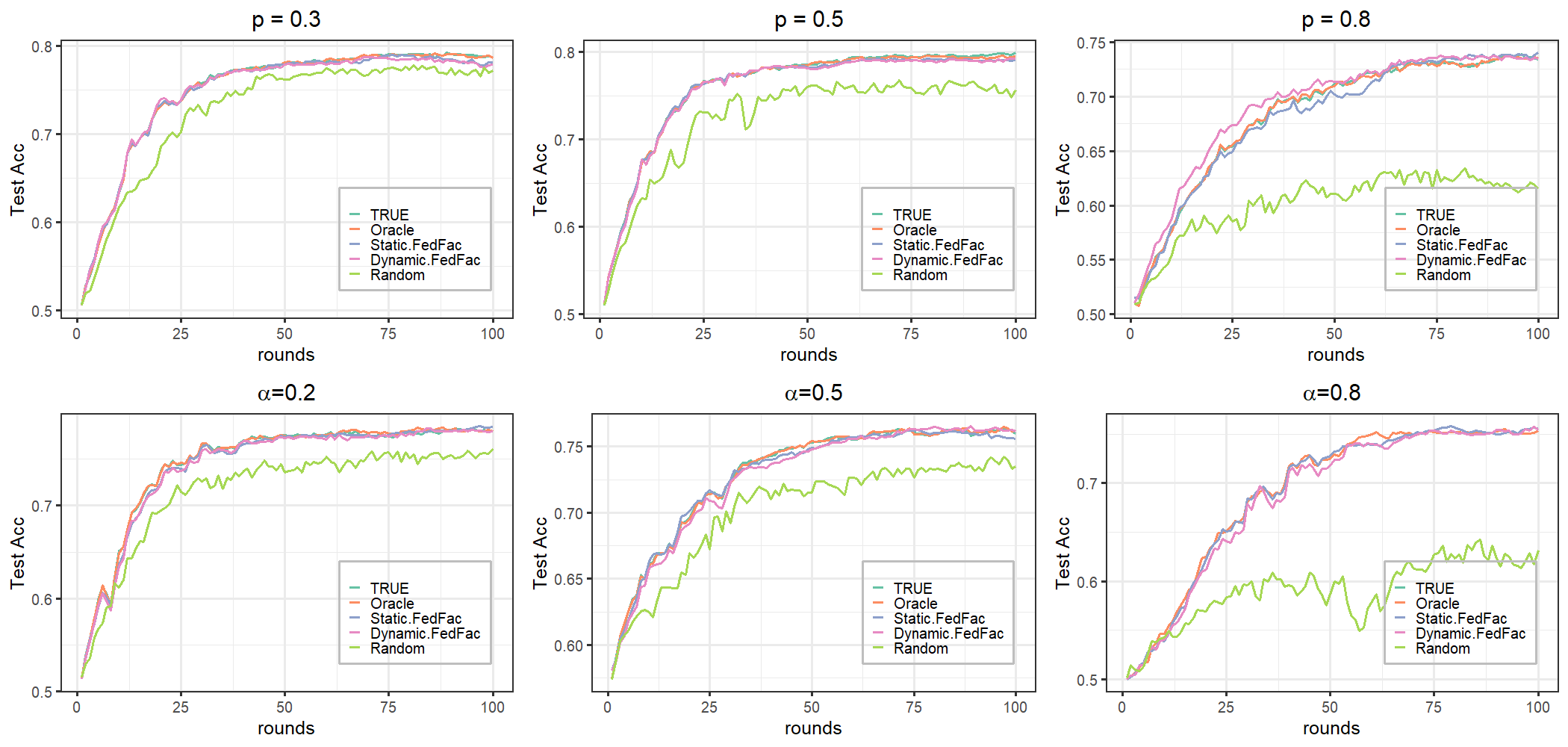} %
	\caption{Test accuracies under different decomposition methods. We vary $p$ in the top three figures by fixing $\alpha=0.4$; and vary $\alpha$ in the bottom three figures by fixing $p=0.5$. The smaller the value of $p$, the more heterogeneous of model. The larger the value of $\alpha$, the more identical of covariates.}
	\label{simu1}
\end{figure}

Recall that by using the static method, the decomposition of client-shared elements and client-specific elements are fixed during the model training procedure. However, the dynamic method behaves more flexible which allows the decomposition to change. Then one question arises: whether the dynamic method can reach convergence? In other words, whether the client-shared elements and client-specific elements can stay unchanged in the end? To investigate this question, we compute the proportion of neurons which do not change states between consecutive rounds when using dynamic FedFac. Figure~\ref{simu2} shows the corresponding results under different experimental settings. As shown, by using dynamic FedFac, nearly all neurons will reach their final stable states during the model training procedure. These findings demonstrate the effectiveness and feasibility of dynamic FedFac in parameter decomposition. In conclusion, dynamic FedFac utilizes more information than static FedFac to decompose the hidden elements and behaves more flexible. It can also reach a static decomposition situation in the model training procedure. Therefore, compared with static FedFac, we recommend to use dynamic FedFac in real applications. Our experiments in real datasets also reveal that dynamic FedFac achieves better prediction results than static FedFac, the details of which can be found in the next section.

\begin{figure}[h]
	\centering
	\includegraphics[width=0.9\textwidth]{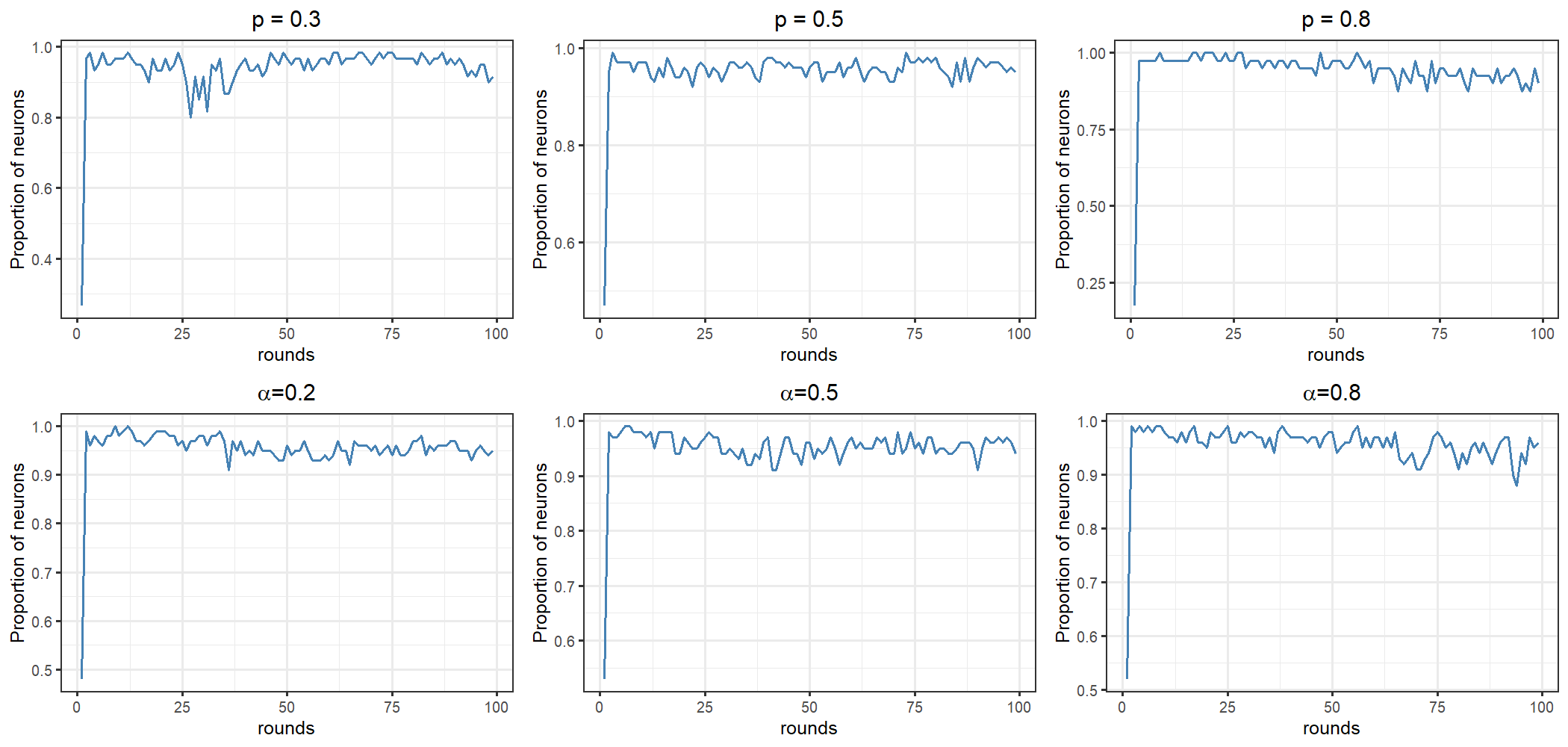} %
	\caption{The proportion of neurons which do not change states between consecutive rounds using dynamic FedFac.}
	\label{simu2}
\end{figure}

\section{Experiments of FedFac on Real Data}\label{experiments}


\subsection{Experimental setup}\label{subsection4}

We conduct various experiments on a wide range of learning tasks and neural networks. Below, we first introduce the datasets, models, and FL approaches used for comparison.

\paragraph{Datasets.} We test the performance of FedFac on four real datasets involving three machine learning problems: (1) FEMNIST \citep{caldas2018leaf}, which is used for handwritten digit recognition, (2) Shakespeare, which is used for next-character prediction \citep{caldas2018leaf}, and (3) CIFAR10 and CIFAR100, which are used for complex image classification \citep{krizhevsky2009learning}. Among these datasets, FEMNIST is a commonly used dataset in federated learning, whose data points on each client are generated from a specific writer. The Shakespeare dataset is naturally heterogeneous with different clients representing different speaking roles \citep{mcmahan2017communication}. For CIFAR10 and CIFAR100, we follow the common practice and consider different levels of label heterogeneity. For CIFAR10, we follow the work \citep{hsu2019measuring} and assign data points of the same label to different clients according to a symmetric Dirichlet distribution with parameter $\pi$. The Dirichlet parameter $\pi$ indicates the non-IID degree with smaller $\pi$ representing more highly skewed local distributions. For CIFAR100, we adopt the pathological non-IID partition as in \citep{mcmahan2017communication}, and the heterogeneity degree is controlled by varying the number of classes $S$ per client.

\paragraph{Models.} Based on the four datasets, we consider a range of classical neural network models, including ResNet, VGG, and LSTM. The detailed description of neural network models and the corresponding used datasets are summarized in App.D. For simplicity, all models are trained with the ADAM optimizer \citep{kingma2014adam} with an initial learning rate of $10^{-3}$ and $\beta = (0.9, 0.999)$. Table \ref{table3} summarizes the datasets and the corresponding used deep neural networks.

\begin{table}[h]
	\centering
	\footnotesize
	\caption{Summary of datasets and models.}
	\label{table3}
	{\tabcolsep=1.5pt
		\begin{tabular}{@{}ccccc@{}}
			\toprule
			Dataset          & Task                          & Clients & Total samples & Model             \\ \midrule
			FEMNIST          & handwritten digit recognition & 1079    & 245,114        &  2-layer CNN + 2-layer FC \\
			Shakespeare      & Next-Character Prediction     & 778     & 4,101,468       &  2-layer LSTM + 2-layer FC \\
			CIFAR10          & Image classification          & 100     & 60,000         & ResNet-18                 \\
			CIFAR100         & Image classification         & 100     & 60,000         & VGG-16                    \\
			\bottomrule
	\end{tabular}}
\end{table}

\paragraph{Competitors.} We compare FedFac against two global FL methods, including: (1) FedAvg \citep{mcmahan2017communication}, (2) FedProx \citep{li2020federated}, as well as four personalized FL methods, including: (1) FedPer \citep{arivazhagan2019federated}, (2) LG-FedAvg \citep{liang2020think}, (3) FedEM \citep{marfoq2021federated}, and (4) FedRep \citep{collins2021exploiting}. Specifically, the penalization parameter $\mu$ in FedProx is tuned by grid search on the set $\{1, 0.5, 0.1, 0.01, 0.001\}$. For FedPer and FedRep, we keep the last layer of the DNN models private. For LG-FedAvg, the number of DNN layers spread across the local and global models is tuned from $1$ to $L-1$. As for FedEM, we set the number of components to be $3$ as suggested by the authors. For all the FL methods, we randomly split the dataset in each client to be training (80\%) and testing (20\%). All experiments are examined in the partial participation case with $C$ denoting the client participation rate. The number of local epochs between each two communication rounds is tuned on the set $\{1, 5, 20\}$.

\paragraph{Tuning Parameter Setup} The FedFac method involves three tuning parameters: the layers $\Lambda$ to be split, the personalization threshold $\tau_l$, and the variability threshold $\kappa_l$ used in estimating $\mathbf{A}_l$. We tune the parameter $\Lambda$ through the set $\{1, \ldots, L-1\}$. As for $\tau_l$, define $\boldsymbol{\nu}_l = \{\nu_{l1},\cdots,\nu_{ld_l}\}$. Then the tuning set of $\tau_l$ contains the 25\%, 50\%, 75\% quantiles of $\boldsymbol{\nu}_l$, along with the $\max(\boldsymbol{\nu}_l)+0.1$, which represents all  parameters are client-specific and denoted by $+\infty$. Note that a larger value of $\tau_l$ indicates more personalized parameters. Then by considering different values of $\tau_l$, the FedFac method can cover FedAvg as well as many other split-personalization methods (e.g., FedPer and LG-FedAvg). This implies FedFac is a more general FL framework. For the parameter $\kappa_l$, we consider a tuning set $\{0.5, 0.75, 0.85, 0.9, 0.95 \}$. All the parameters are tuned by grid search within their tuning sets.

\subsection{Comparison results}

We evaluate the model performance on the local test dataset unseen at training for all FL approaches. Specifically, the averaged weighted accuracy across all clients in the FL system is computed by using weights proportional to the local dataset sizes. Table~\ref{table1} summarizes the best averaged accuracy measured across 10 consecutive rounds after convergence is achieved setting $C=0.1$. It is observed that, FedFac significantly outperforms all competing methods in almost all settings. Dynamic FedFac always performs better than its static counterpart, suggesting the benefit of adjusting the initial decomposition. Besides, we find the decomposition of parameters would reach a stable state in dynamic FedFac. Thus we use dynamic FedFac for the following analysis and omit ``dynamic'' to save space. The results on CIFAR10 and CIFAR100 show that, FedFac is robust to different levels of heterogeneity. In summary, these results demonstrate the benefits and applicability of FedFac.


\begin{table*}[h]
	\centering
	\caption{Average test accuracies on different datasets. The decomposed layer $l$ using dynamic FedFac is also reported in the last column. }
	\scriptsize
	\tabcolsep=0.1cm
	\label{table1}
	\begin{tabular}{@{}ccccccccccc@{}}
		\toprule
		\multirow{2}{*}{Dataset}                   & \multirow{2}{*}{Setting} & \multirow{2}{*}{FedAvg} & \multirow{2}{*}{FedProx}      & \multirow{2}{*}{FedPer} & \multirow{2}{*}{LG-FedAvg} & \multirow{2}{*}{FedEM} & \multirow{2}{*}{FedRep} & \multicolumn{2}{c}{FedFac} & \multirow{2}{*}{$\Lambda$} \\
\cline{9-10}
&&&&&&&&Static &Dynamic & \\
\midrule
		\multirow{3}{*}{CIFAR10} & $\pi = 0.1$    & 0.8033 & 0.8058   & 0.7624 & 0.7710    & 0.5722  & 0.6694 & \textbf{0.8307} & \textbf{0.8307} & \{16,17\} \\
		& $\pi = 0.5$     & 0.7039 & 0.7196   & 0.7130 & 0.5544    & 0.5708 & 0.3932 & 0.7483 &\textbf{0.7487} & \{17\} \\
		& $\pi = 1$       & 0.6849 & 0.7007   & 0.6870 & 0.4702    & 0.5331 & 0.3072 & 0.7047 & \textbf{0.7153} & \{17\} \\ \midrule
		\multirow{3}{*}{CIFAR100} & $S = 5$       &   0.0710     &  0.2093       &    0.6773    & 0.6586   &   0.5224 & 0.2153 & \textbf{0.7138} & 0.7136  & \{15\}  \\
		& $S = 10$      & 0.3011 & 0.3313   & 0.6643 & 0.2435    & 0.5189 & 0.1725 & 0.6897 &\textbf{0.6932} & \{15\} \\
		& $S = 30$      & 0.4513 & 0.4658   & 0.5994 & 0.2243    & 0.6183 & 0.1107 & 0.6107 & \textbf{0.6275} & \{15\} \\ \midrule
		FEMNIST         & ----        & 0.7680 & 0.8107   & 0.7123 &  0.7653   & 0.7949 & 0.6965 & 0.8139 & \textbf{0.8301} & \{2\} \\ \midrule
		Shakespeare     & ----        & 0.4996 & 0.4992    & 0.5014 &  0.3636   & 0.3533 & 0.4095 & 0.4983 & \textbf{0.5029} & \{3\} \\ \bottomrule
	\end{tabular}
\end{table*}

We further investigate the local performance of different clients, whose detailed results are shown in Figure \ref{cifar}. It is clear that, FedFac has gained improvement in terms of the local accuracy for most clients. In addition, note that the FedFac method can achieve relatively small variance of client-side accuracies. It implies the advantage of FedFac in reducing the heterogeneity of model performance across different clients. In other words, the fairness across clients in terms of local performances in different clients can be improved by FedFac.

\begin{figure}[h]
	\centering
	\includegraphics[width=1\textwidth]{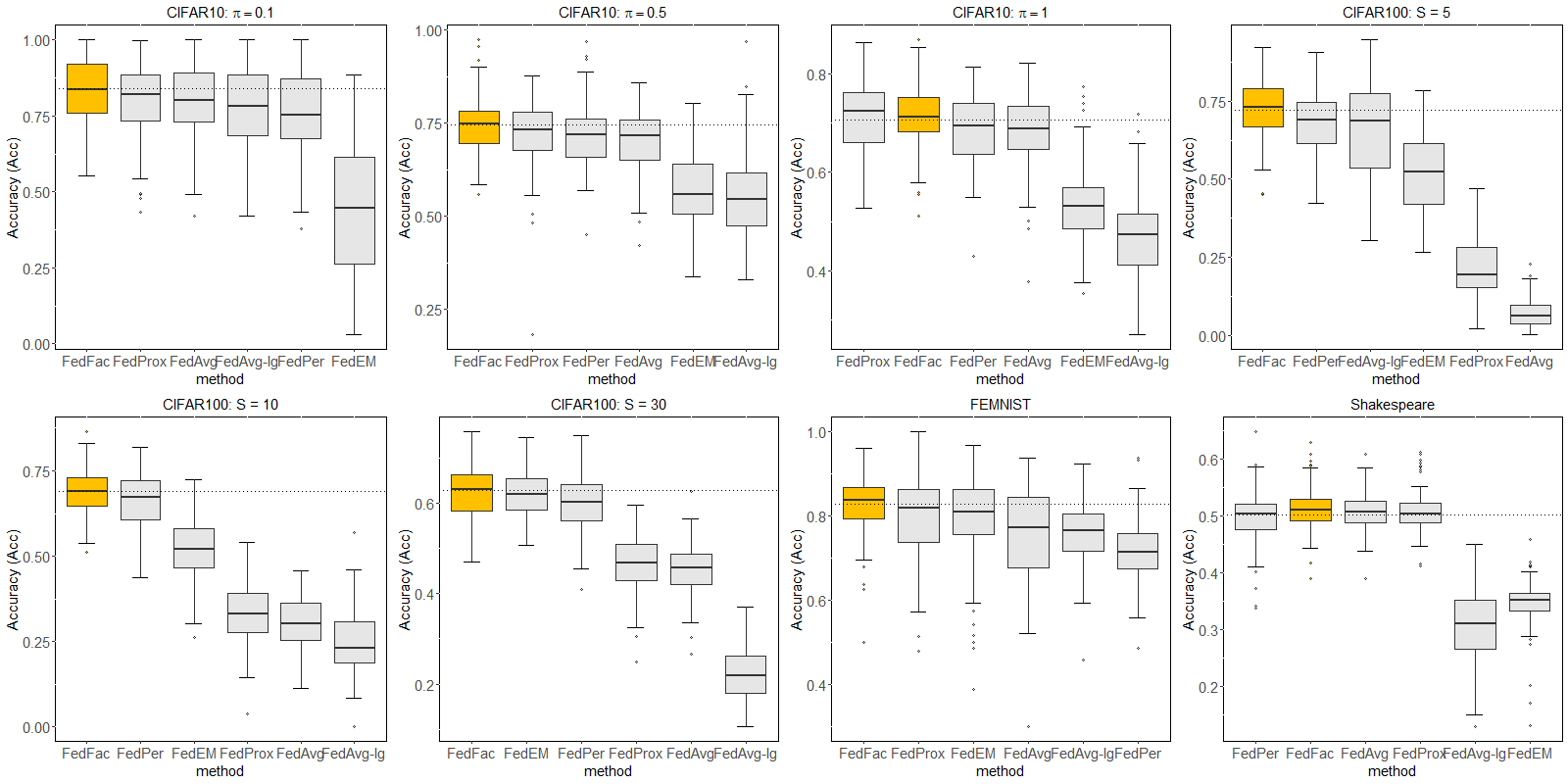} %
	\caption{Local performance across all clients on varied datasets. Different models are ordered by their median local performance. The orange box represents our proposed FedFac method.}
	\label{cifar}
\end{figure}

Last, we focus on the prediction performance on new clients. To this end, we randomly select 10\% of the total clients as new clients, and prevent them from model training. Then we apply the two solutions (LocalTrain and Ensemble) discussed in Section 3.2 to predict labels in the new clients. As for the competitors, we apply their original generalization methods for new clients. We do not include FedPer because it did not mention the generalization method. For FedAvg and FedProx, we just apply the estimated global models on the new clients. The obtained global models with fine-tune are also considered as baseline, which are denoted by FedAvg+ and FedProx+.
The method LG-FedAvg applies a similar ensemble approach for new clients; while FedEM applies a similar LocalTrain approach.
The experiment is conducted on CIFAR10 for illustration purpose.
Table~\ref{table-g} shows the prediction performance achieved by FedFac and its competitors. As shown, the FedFac method with Ensemble always performs better than that with LocalTrain. It also outperforms all its competitors in the cases $\pi = 0.5$ and 1. In the extreme heterogeneity case with $\pi = 0.1$, FedFac with Ensemble can still achieve similar performances with the fine-tune methods FedAvg+ and FedProx+.


\begin{table}[h]
	\scriptsize
	\caption{The test accuracies of new clients on CIFAR10. }
	\label{table-g}
	\tabcolsep=0.1cm
	\begin{center}
		\newcommand{\tabincell}[2]{\begin{tabular}{@{}#1@{}}#2\end{tabular}}
		\begin{tabular}{l|ccccccccc}
			\toprule
			\multirow{2}{*}{Setting}     & \multirow{2}{*}{FedAvg} & \multirow{2}{*}{FedAvg+} & \multirow{2}{*}{FedProx} & \multirow{2}{*}{FedProx+} & \multirow{2}{*}{LG-FedAvg} & \multirow{2}{*}{FedEM} & \multirow{2}{*}{FedRep} & \multicolumn{2}{c}{FedFac} \\
\cline{9-10}
&&&&&&&&LocalTrain &Ensemble\\
\midrule
			$\pi = 0.1$ & 0.4004 & 0.6500 & 0.2813 & \textbf{0.6520} & 0.2046  & 0.3701 & 0.3970 & 0.6113  & 0.6410 \\
			$\pi = 0.5$  & 0.5953 & 0.6604 & 0.4948 & 0.6799  & 0.3109  & 0.5807  & 0.3319 & 0.5750 & \textbf{0.7154} \\
			$\pi = 1$  & 0.6847 & 0.6953 & 0.6636 & 0.7003 & 0.3770   & 0.5193 & 0.2660 & 0.7034  & \textbf{0.7160}  \\ \bottomrule
		\end{tabular}
	\end{center}
	\vskip -0.1in
\end{table}

\subsection{The tuning procedure of hyper-parameters}\label{more_para_tune}
Recall that we have three hyper-parameters $(\Lambda,\tau_l,\kappa_l)$. Our preliminary exploration shows $\Lambda$ has the largest influence on the classification performance of FedFac, which is followed by $\tau_l$ and $\kappa_l$. Therefore, to avoid time consuming in grid search, we suggest the following parameter tuning procedure. That is, tune $\Lambda$ first, then tune $\tau_l$ with the optimal $\Lambda$ fixed, and finally tune $\kappa_l$ with the optimal $\Lambda$ and $\tau_l$ fixed. To illustrate this tuning procedure, we take CIFAR100 as an example, which is estimated by VGG-16.

Table \ref{t:tune1} summarizes the test accuracies under different $\Lambda$, with $\tau_l=50$ and $\kappa_l=0.85$ fixed. We find that, the largest $\Lambda$ results in the best test accuracy for different setups of $S$. Recall $\Lambda$ determines the layers to conduct decomposition for client-shared parameters and client-specific parameters. This finding suggests that, for more complicated neural networks dealing with challenging tasks (i.e., CIFAR100 with VGG-16), conducting decomposition on deeper layers results in better model performance than that on shallow layers. This is because the deeper layers usually express high-level representations, which may behave more helpful to complicated tasks. On the contrary, for simple DNN structures tackling easier tasks (i.e., the handwritten digit recognition), our results show that, shallow layers are more preferable and should be taken into consideration.

By Table \ref{t:tune1}, we have chosen the best $\Lambda$ for CIFAR100, i.e., $\Lambda=\{15\}$. In the next step, we fix $\Lambda=\{15\},\kappa_l=0.85$, and then select the optimal $\tau_l$. The corresponding results are shown in Table \ref{tau_l}. We find that, the optimal $\tau_l$ is $95\%$ for $S=5$ and $S=10$. For $S=30$, the optimal $\tau_l$ is the 90\% quantile of $\boldsymbol{\nu}_l$. Recall $S$ controls the heterogeneity level of data. The smaller the $S$, the larger heterogeneity of the data. Therefor, these results suggest that, the optimal value of $\tau_l$ is related with the heterogeneity level of data. Larger $\tau_l$ is more beneficial when data have higher level of heterogeneity among different clients. This finding also confirms our intuitive understanding that, high heterogeneity inherently encourages more personalization and less sharing. Finally, with the optimal $\Lambda$ and $\tau_l$ fixed, we can select $\kappa_l$. Note that $\kappa_l$ is used to determine the number of common factors in factor analysis. Table \ref{kappa_l} shows the tuning results of $\kappa_l$ for the three cases.

\begin{table}[h]
	\caption{Tuning results of $\Lambda$ on CIFAR100 with $\tau_l = 50, \kappa_l = 0.85$.}
	\label{t:tune1}
	\scriptsize
	\begin{center}
		\begin{tabular}{@{}c|cccccccc@{}}
			\toprule
			$\Lambda$    & \{1\}       & \{3\}       & \{4\}  & \{5\}  & \{6\}  & \{7\}  & \{8\}  \\ \midrule
			$S=5$  & 0.1043  & 0.1178   &  0.0938  &  0.1075  & 0.1078   &  0.1397  & 0.1773   \\
			$S=10$ & 0.2967  & 0.3506  & 0.3678   & 0.4050   & 0.4219   & 0.4180   & 0.4986   \\
			$S=30$ & 0.4433  & 0.4601  & 0.4618   & 0.4580   & 0.4683   & 0.5032   & 0.5219   \\ \hline
			$\Lambda$    & \{9\}      & \{10\}    & \{11\}   & \{12\}   & \{13\}   & \{14\}       & \{15\} \\ \midrule
			$S=5$  & 0.1418  & 0.1516  & 0.2514   & 0.1563   & 0.0718   & 0.1876   & 0.6822   \\
			$S=10$ & 0.5258  & 0.5743  & 0.5167   & 0.3887   & 0.2275   & 0.3912   & 0.6561   \\
			$S=30$ & 0.5412  & 0.5535  & 0.5711   & 0.5725   & 0.3943   & 0.5615   & 0.6100 \\ \bottomrule
		\end{tabular}
	\end{center}
\end{table}

\begin{table}[h]
	\caption{Tuning results of $\tau_l$ on CIFAR100 with $\Lambda=\{15\}, \kappa_l = 0.85$.}
	\label{tau_l}
	\scriptsize
	\begin{center}
		\begin{tabular}{@{}l|ccccc@{}}
			\toprule
			$\tau_l$	& 25\%     & 50\%     & 75\%     & 90\%  & 95\% \\ \midrule
			$S=5$  & 0.6020 & 0.6792 & 0.6883 & 0.6940 & \textbf{0.7136}   \\
			$S=10$ & 0.2739 & 0.4254 & 0.4814 & 0.6788 & \textbf{0.6932}   \\
			$S=30$ & 0.5747 & 0.6066 & 0.6189 & \textbf{0.6260} & 0.6072   \\ \bottomrule
		\end{tabular}
	\end{center}
\end{table}

\begin{table}[h]
	\caption{Tuning results of $\kappa_l$ on CIFAR100 with $\Lambda=\{15\}$ and according optimal $\tau_l$ in Table~\ref{tau_l}.}
	\label{kappa_l}
	\scriptsize
	\begin{center}
		\begin{tabular}{@{}l|ccccc@{}}
			\toprule
			$\kappa_l$	& 0.5     & 0.75     & 0.85     & 0.9  & 0.95 \\ \midrule
			$S=5$  & 0.6921 & 0.6973  & 0.7031 & \textbf{0.7136} & 0.7084   \\
			$S=10$ & 0.6846 & \textbf{0.6932} & 0.6778 & 0.6847 & 0.6893   \\
			$S=30$ & 0.6145 & \textbf{0.6260} & 0.6134 & 0.6090 & 0.6103   \\ \bottomrule
		\end{tabular}
	\end{center}
\end{table}

\newpage
\subsection{The computational issue}\label{E3}
We first investigate the convergence rate of FedFac using the dynamic method. Figure~\ref{loss} presents the training loss curves of FedFac and FedAvg in different datasets. Compared with FedAvg, we find the loss of FedFac goes down faster in all datasets with different heterogeneity settings, indicating that FedFac has a faster convergence rate than FedAvg. This result is consistent with our theoretical analysis in Theorem~\ref{theorem2}. Moreover, compared to FedAvg, FedFac behaves smoother and more stable during the learning process.
We then focus on the influence of heterogeneity levels. We can find that, the higher heterogeneity of data
across clients, the larger difference between the loss curves of FedFac and FedAvg. This finding suggests that, the convergence rate of FedFac is much faster than that of FedAvg in more heterogeneous situations. In particular, the two loss curves nearly coincide for nearly IID datasets  (i.e., CIFAR10 with $\pi = 1$ and Shakespeare), which implies the convergence rate of FedFac is almost the same as FedAvg when data have nearly no heterogeneity.

\begin{figure}[h]
	\centering
	\includegraphics[width=1\textwidth]{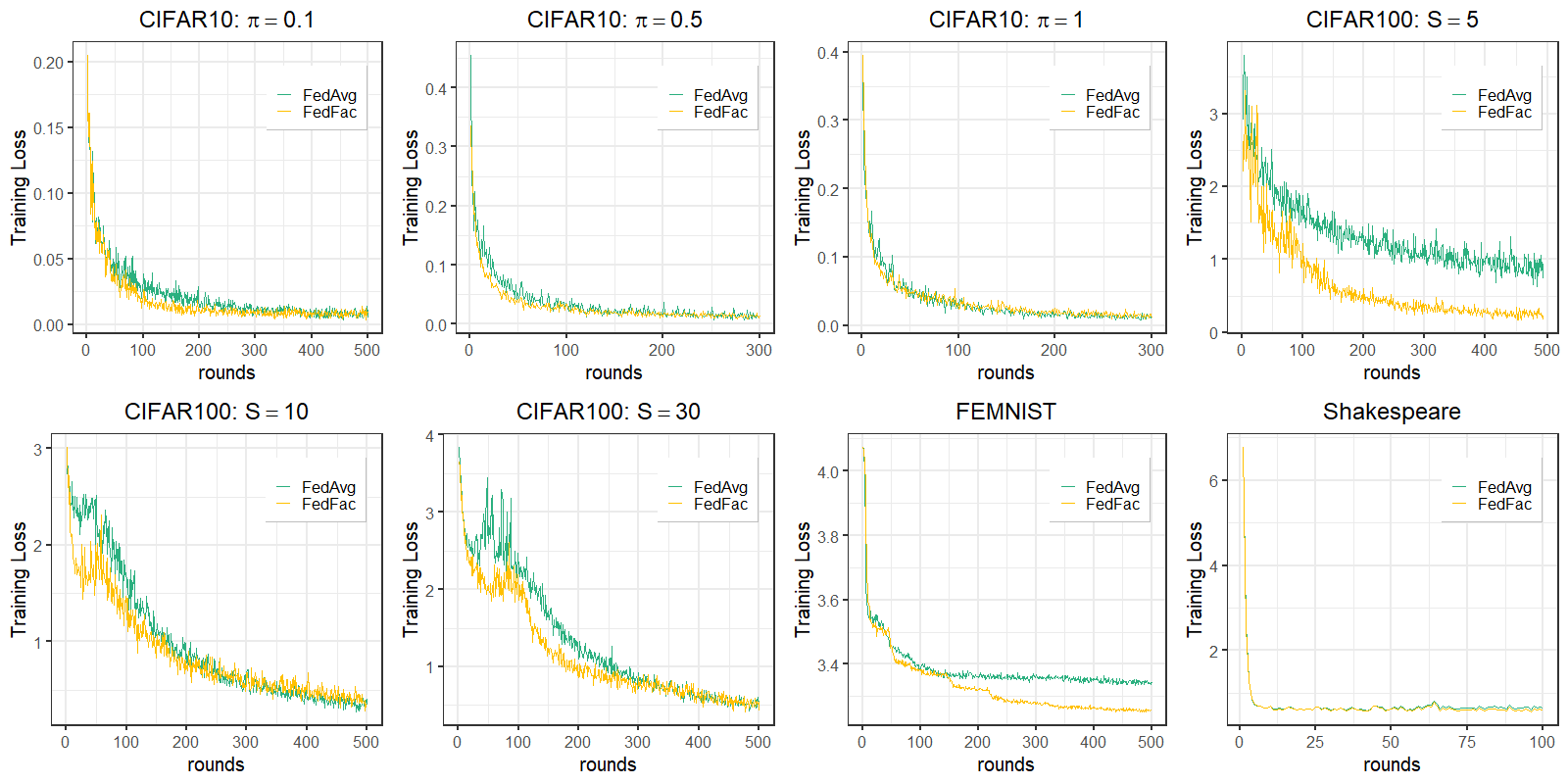}
	\caption{The curves of training loss obtained by FedFac and FedAvg on different datasets.}
	\label{loss}
\end{figure}

We also explore the computational complexity of FedFac. To this end, we take the running time consumed by FedAvg in the training process as baseline. Then we report the ratio between the running time consumed by a particular method (e.g., FedFac) with FedAvg. Table \ref{tableE4} presents the detailed results. As shown, FedFac indeed consumes more time than FedAvg because of the implementation of factor analysis. However, compared with other competitors addressing data heterogeneity, FedFac is still competitive.

\begin{table*}[h]
	\centering
	\caption{The ratio of training time of all methods relative to that of FedAvg.}
	\scriptsize
	\tabcolsep=0.1cm
	\label{tableE4}
	\begin{tabular}{@{}cccccccc@{}}
		\toprule
		Dataset    & Setting & FedProx      & FedPer & LG-FedAvg & FedEM & FedRep & FedFac  \\ \midrule
		\multirow{3}{*}{CIFAR10} & $\pi = 0.1$   & $4.2073$   & $1.0038$ & $1.0032$  & $2.8738$  & $0.6683$ & $0.8559$  \\
		& $\pi = 0.5$      & $4.7352$   & $1.1357$ & $1.1337$    & $2.8508$ & $0.8679$ & $1.1798$ \\
		& $\pi = 1$       & $4.5255$   & $0.9828$ & $0.9996$    & $2.8091$ & $0.7292$ & $1.0845$ \\ \midrule
		\multirow{3}{*}{CIFAR100} & $S = 5$  & $1.2945$  & $0.6967$  & $0.6884$  & $12.2676$ & $2.7761$ & $3.3879$  \\
		& $S = 10$       & $1.3165$   & $0.9118$ & $0.7055$    & $9.3704$ & $1.8907$ & $3.2842$  \\
		& $S = 30$       & $1.5052$   & $0.7143$ & $0.5686$  & $8.1725$ & $1.9346$ & $2.4466$ \\ \midrule
		FEMNIST         & ----  & $2.2718$   & $1.1752$ & $1.1045$ & $9.3726$ & $0.6964$ & $1.2147$ \\ \midrule
		Shakespeare     & ----  & $3.5878$ & $0.9052$ & $0.9460$ & $10.4129$ & $0.7915$ & $1.2856$\\ \bottomrule
	\end{tabular}
\end{table*}

\subsection{Ablation studies}
We conduct ablation studies from two perspectives. First, we aim to investigate the effect of client-shared parameters and client-specific parameters. To this end, we mask each group of parameters in the $l$-th layer when updating the local models, and then record the corresponding performance. Here, the \emph{mask} operation means replacing the targeted group of parameters by random values. For fair comparison, we set $\tau_l = 50\%$, which means the two groups have equal size of parameters. Panel A in Table~\ref{table4} presents the performance when masking the shared parameters or personalized parameters in different experimental settings. As shown, in most settings especially the heterogeneity level is high, masking personalized parameters would lead to worse classification performance than masking the shared parameters. This finding demonstrates the usefulness of personalized parameters. Second, we aim to investigate the effect of decomposition by factor analysis. To this end, we randomly select the shared and personalized parameters for the layers to be split while keeping the other experiment settings consistent with FedFac. The results are shown in Panel B of Table~\ref{table4}. For clear comparison, the corresponding results by FedFac in Table ~\ref{table4} are also reported. As shown, FedFac can always achieve better performance than random split, which verifies the usefulness of factor analysis.

\begin{table}[h]
	\centering
	\caption{The performance evaluated by Acc of two settings in ablation studies. }
	\scriptsize
	\tabcolsep=0.1cm
	\label{table4}
	\newcommand{\tabincell}[2]{\begin{tabular}{@{}#1@{}}#2\end{tabular}}
	\begin{tabular}{@{}ccccccccc@{}}
		\toprule
		Setting & \tabincell{c}{CIFAR10 \\$\pi = 0.1$} & \tabincell{c}{CIFAR10 \\ $\pi = 0.5$} & \tabincell{c}{CIFAR10 \\ $\pi = 1$} & \tabincell{c}{CIFAR100 \\$S = 5$} & \tabincell{c}{CIFAR100 \\$S = 10$} & \tabincell{c}{CIFAR100 \\$S = 30$} & FEMNIST & Shakespeare \\ \midrule
		&\multicolumn{8}{c}{\emph{Panel A: Client-shared V.S. Client-specific}}\\
		\cmidrule{2-9}
		\tabincell{c}{mask-shared} & 0.8154 & 0.7417 & 0.7092 & 0.6794  & 0.6618  & 0.6052  & 0.8256 & 0.5010 \\
		\tabincell{c}{mask-personalized} & 0.7225 & 0.7043 & 0.7037 & 0.6578 & 0.6448 &  0.5846 & 0.4793 & 0.4993 \\
		\midrule
		&\multicolumn{8}{c}{\emph{Panel B: Factor Analysis V.S. Random Split}}\\
		\cmidrule{2-9}
		factor analysis & 0.8307 & 0.7487 & 0.7153 & 0.7138 & 0.6932 &  0.6275 & 0.8301 & 0.5029\\
		random split & 0.8140 & 0.7251 & 0.7037 & 0.6831 & 0.6808 &  0.6060 & 0.8113 & 0.4990\\
		\bottomrule
	\end{tabular}
\end{table}

\subsection{Interpretability}
We visualize the featuremaps generated by the personalized and shared filters obtained by FedFac. For comparison purpose, we also visualize the representations of corresponding filters in FedAvg.
For illustration, we take the fifth convolutional layer (the first convolutional module) of ResNet-18 trained on CIFAR10 with $\pi = 0.5$ for example, since the feature maps outputted by shallow layers are less abstract compared with deeper layers (which usually output pigment matrix).
We set $\tau_5 = 50\%$ so that half of channels in this layer are personalized.
As shown in Figure~\ref{interpre2}, the personalized filters in FedFac can extract more clear and specific features for the original \emph{Image A}, while the corresponding shared filters in FedAvg generate much more noise in feature representations. It again demonstrates the advantage of personalized filters in FedFac. The shared filters in FedFac behave similarly with those in FedAvg. In addition, we find the two types of filters (personalized v.s. shared) in FedFac generate quite different representations. The personalized filters probably more focus on extracting target-specific features, whereas the shared filters tend to summarize the overall features of the original figure. These results justify our motivation to separate the client-shared representations and client-specific representations.

\begin{figure*}[h]
	\centering
	\includegraphics[width=0.95\textwidth]{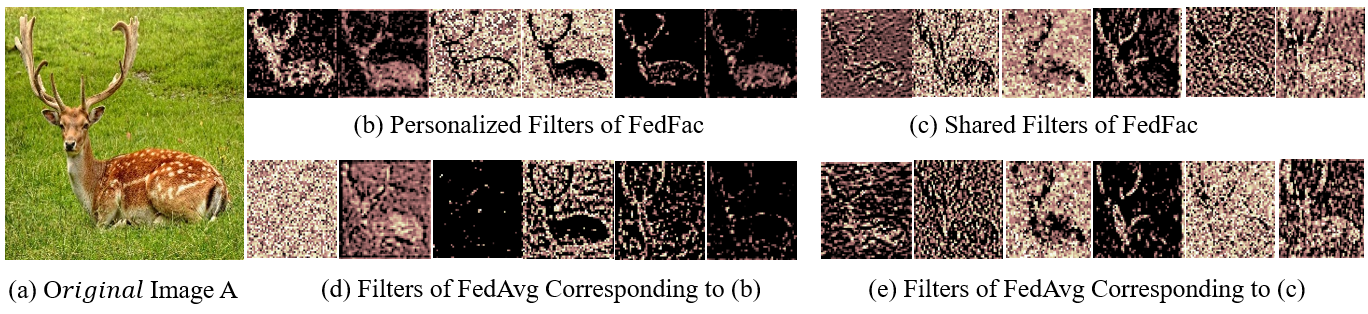} 
	\caption{Representations of one image generated by the fifth convolution layers of ResNet-18.}
	\label{interpre2}
\end{figure*}

\section{Conclusion and Discussion} \label{section5}
In this work, we propose a novel personalization approach called \emph{FedSplit} to tackle the statistical heterogeneity problem in federated learning. We are motivated by a common finding that, data in different clients could contain both common knowledge and personalized knowledge. Thus the hidden elements within the same layer should be also decomposed into client-shared ones and client-specific ones. With this decomposition, the parameters corresponding to the client-shared group are updated by all clients, while the parameters corresponding to the client-specific group are only updated locally and not averaged by the server. By this way, the resulting models are more customized for individual clients. We demonstrate FedSplit enjoys a faster convergence speed than the standard FedAvg method. The generalization bound of FedSplit is also analyzed. To practically decompose the two groups of information, the factor analysis technique is applied. This leads to the FedFac method, which is a practically implemented version of FedSplit. We demonstrate by simulation studies that, using factor analysis can well recover the underlying shared/personalized decomposition. Extensive experiments also show that, FedFac can improve model performance than various state-of-the-art federated learning algorithms.

The FedFac method also has some limitations, which inspire further study in the future.
First, some automatical hyper-parameter tuning strategy such as choosing the network layer
for decomposition is worth of consideration. Second, some privacy-protecting techniques (e.g., differential privacy) can be leveraged to protect the proposed method from potential disclosure of sensitive information. Last, our theoretical analysis is applicable for the static FedFac method. The theoretical properties of using dynamic FedFac for decomposition are worth of further investigation.

\acks{Yang Li (yang.li@ruc.edu.cn) is the corresponding author.} This work is supported by National Natural Science Foundation of China (No.72371241, 72001205, 72271237), Chinese National Statistical Science Research Project (2022LD06), the MOE Project of Key Research Institute of Humanities and Social Sciences (22JJD910001).

\section*{Conflict of Interest}
The authors declare they have no conflict of interest.

\vskip 0.2in
\bibliography{jmlr_main}

\end{document}